\documentclass[journal]{IEEEtran}
\usepackage{graphicx}
\usepackage{cite}
\usepackage{picinpar}
\usepackage{amsmath,bm}
\usepackage{url}
\usepackage[latin1]{inputenc}
\usepackage{colortbl}
\usepackage{soul}
\usepackage{multirow}
\usepackage{pifont}
\usepackage{color}
\usepackage{alltt}
\usepackage[hidelinks]{hyperref}
\usepackage{enumerate}
\usepackage{siunitx}
\usepackage{breakurl}
\usepackage{epstopdf}
\usepackage{pbox}
\usepackage{indentfirst}
\usepackage{booktabs}
\usepackage{subfigure}
\usepackage{float}
\usepackage{threeparttable}
\usepackage[numbers,sort&compress]{natbib}
\usepackage{algorithmic}
\usepackage{algorithm}
\makeatletter
\newcommand{\removelatexerror}{\let\@latex@error\@gobble}
\makeatother

\begin{document}

\title{Wheel-INS2: Multiple MEMS IMU-based Dead Reckoning System with Different Configurations for Wheeled Robots}

\author{Yibin~Wu, Jian~Kuang, and~Xiaoji~Niu
\thanks{This work was funded in part by the National Key Research and Development Program of China (No. 2016YFB0501800 and No. 2016YFB0502202).\textit{(Corresponding Author: Jian Kuang.)}}
\thanks{The authors are with the GNSS Research Center, Wuhan University, Wuhan, China. \{ybwu, kuang, xjniu\}@whu.edu.cn.}}

\maketitle

\begin{abstract}
A reliable self-contained navigation system is essential for autonomous vehicles. Based on our previous study on Wheel-INS \cite{niu2019}, a wheel-mounted inertial measurement unit (Wheel-IMU)-based dead reckoning (DR) system, in this paper, we propose a multiple IMUs-based DR solution for the wheeled robots. The IMUs are mounted at different places on the wheeled vehicles to acquire various dynamic information. In particular, at least one IMU has to be mounted at the wheel to measure the wheel velocity and take advantage of the rotation modulation. The system is implemented through a distributed extended Kalman filter structure where each subsystem (corresponding to each IMU) retains and updates its own states separately. The relative position constraints between the multiple IMUs are exploited to further limit the error drift and improve the system's robustness. Particularly, we present the DR systems using dual Wheel-IMUs, one Wheel-IMU plus one vehicle body-mounted IMU (Body-IMU), and dual Wheel-IMUs plus one Body-IMU as examples for analysis and comparison. Field tests illustrate that the proposed multi-IMU DR system outperforms the single Wheel-INS in terms of both positioning and heading accuracy. By comparing with the centralized filter, the proposed distributed filter shows unimportant accuracy degradation while holding significant computation efficiency. Moreover, among the three multi-IMU configurations, the one Body-IMU plus one Wheel-IMU design obtains the minimum drift rate. The position drift rates of the three configurations are 0.82\% (dual Wheel-IMUs), 0.69\% (one Body-IMU plus one Wheel-IMU), and 0.73\% (dual Wheel-IMUs plus one Body-IMU), respectively. 

\end{abstract}

\begin{IEEEkeywords}
Wheel-mounted IMU, multi-IMU, dead reckoning, vehicular navigation, wheeled robot.
\end{IEEEkeywords}

\IEEEpeerreviewmaketitle

\section*{Nomenclature}
\begin{enumerate}[a)]
	\item Matrices are denoted as uppercase bold letters
	\item Vectors are denoted as lowercase bold italic letters.
	\item Scalars are denoted as lowercase italic letters.
	\item Coordinate frames involved in the vector transformation are denoted as superscript and subscript. For vectors, the superscript denotes the projected coordinate system.
	\item $\hat{*}$, estimated or computed values.
	\item $\widetilde{*}$, observed or measured values.
	
\end{enumerate}

\section{Introduction}

\IEEEPARstart{I}{nertial} navigation system (INS) is an old but extensively used method \cite{wu2009} to complement the Global Navigation Satellite System (GNSS), so as to provide continuous positioning results for the land vehicles in densely built-up areas. In these complicated scenarios, GNSS positioning deteriorates because of signal jamming and masking, whereas the inertial measurement unit (IMU) can autonomously perceive the full-state motion information of the vehicle in a self-contained manner \cite{aloi2007}. However, the standalone INS drifts rapidly due to inherent sensor errors, especially for the low-end microelectromechanical systems (MEMS) \cite{Naser2010,zhangquan2020}.

To further suppress the error drift of INS during GNSS outages, the odometer and nonholonomic constraint (NHC) have been widely used in vehicular navigation \cite{sukkarieh2000,dissanayake2001}. The NHC refers to that a land vehicle generally cannot move in the directions perpendicular to its forward direction. Basically, odometer and NHC measurements are constructed based on the vehicle velocity information, which would significantly contribute to the stability and accuracy of INS \cite{wu2009,chen2020}.

In the conventional odometer-aided INS (ODO/INS), the IMU is usually placed on the vehicle body or in the trunk. In addition, either an external odometer is installed or the wheel encoder is accessed to obtain the vehicle velocity. However, it is tricky to fuse information from two different types of sensors because of hardware modification, data transfer synchronization, etc \cite{collin2014tvt}. In our previous paper \cite{niu2019}, we proposed to use one sensor modal to implement the same information fusion as ODO/INS, which is called Wheel-INS, a wheel-mounted MEMS IMU (Wheel-IMU)-based DR system. 

In Wheel-INS, the gyro readings of the Wheel-IMU are multiplied by the wheel radius to produce the wheel velocity. Afterward, the wheel speed along with the NHC was integrated with the strapdown INS via the error-state extended Kalman filter (EKF) to suppress the error drift. The installation scheme of the Wheel-IMU, the system characteristics, and the DR error analysis were explained in detail in \cite{niu2019}. In summary, there are two major advantages of Wheel-INS. Firstly, a similar information fusion scheme as ODO/INS can be achieved by only one IMU. Secondly, the continuous rotation of the Wheel-IMU would significantly limit the heading error drift. It was illustrated that the positioning and heading accuracy of Wheel-INS had been respectively improved by 23\% and 15\% over ODO/INS. Moreover, benefiting from the rotation modulation, Wheel-INS showed desirable immunity to the constant gyro bias error. However, due to the lack of the vehicle pitch angle information, we need to assume that the vehicle is moving on the horizontal plane in Wheel-INS. In \cite{wu2020arxiv}, we compared and analyzed the advantages and features of three different measurement models (velocity measurement, displacement increment measurement, and contact point zero-velocity measurement) in Wheel-INS. 



Based on Wheel-INS \cite{niu2019}, it is natural to consider the feasibility of mounting two IMUs on the two non-steering wheels of the land vehicles to obtain double wheel velocity information, so as to improve the DR performance. Additionally, mounting an IMU on the vehicle body (Body-IMU) to measure the vehicle attitude is desirable to extend Wheel-INS from 2D DR to 3D navigation. Furthermore, the constant spatial relationship between multiple IMUs can be exploited as an external measurement to suppress the error drift of INS. In conclusion, \textit{the motivation of this paper is mounting multiple MEMS IMUs to different locations of the wheeled robots to obtain and fuse diverse motion information of the vehicle for a better DR performance}. 

\begin{figure}[htbp]
	\centering
	\includegraphics[width=7cm]{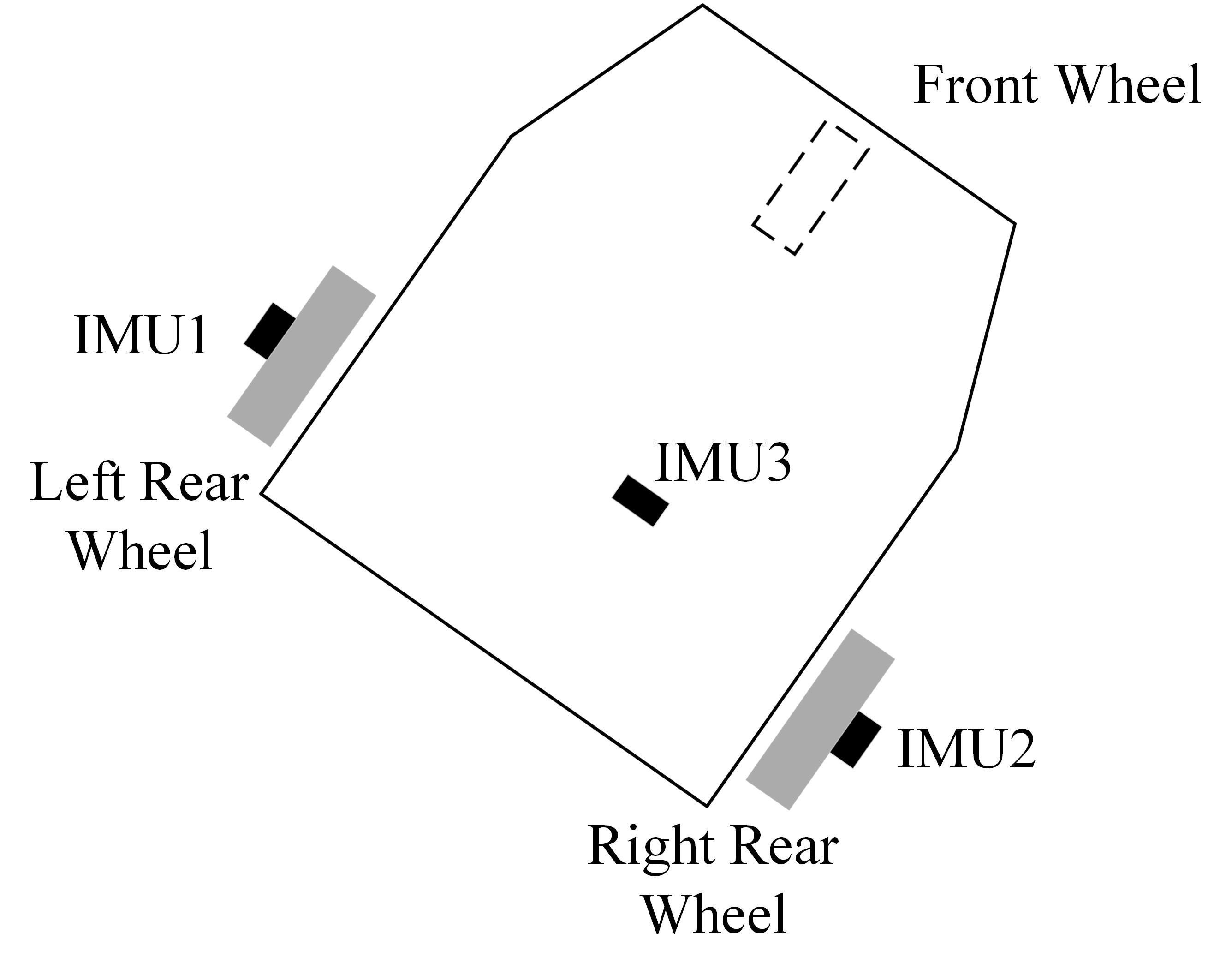}
	\caption{Top view of the locations of multiple IMUs.}
	\label{1}
\end{figure}

The locations of the multiple IMUs are shown in Fig. \ref{1}. According to our previous study \cite{niu2019}, at least one IMU should be mounted on the wheel center, which is indispensable to determining the wheel velocity. The other IMUs can be mounted on other wheels or the vehicle body. In this study, we implement three typical configurations, as shown in Table \ref{Tab1}, to demonstrate and analyze the proposed multi-IMU DR system.  

\begin{table}[h]
	\renewcommand\arraystretch{1.5}
	\centering
	\caption{IMU Configurations of the Three Typical Multi-IMU Systems}
	\label{Tab1}
	\begin{threeparttable}
		\begin{tabular}{p{2.4cm}p{5cm}}
			\toprule
			\multicolumn{1}{c}{Systems} & \multicolumn{1}{c}{IMU Configurations\tnote{*}} \\
			\midrule
			{Dual Wheel-INS} & {Dual Wheel-IMUs (IMU1+IMU2)} 
			\\
			{Body/Wheel-INS} & {One Wheel-IMU plus one Body-IMU (IMU1/IMU2+IMU3)} 
			\\
			{Triple INS} & {Dual Wheel-IMUs plus one
				Body-IMU (IMU1+IMU2+IMU3)} 
			\\
			\bottomrule
		\end{tabular}
		\begin{tablenotes}   
			\footnotesize            
			\item[*]The locations of IMU1, IMU2, and IMU3 are shown in Fig. \ref{1}.      
		\end{tablenotes}
	\end{threeparttable}
\end{table}


The remaining content of this paper is organized as follows. Relevant literature focusing on wheel-mounted inertial sensors and multi-IMU fusion is discussed in Section \uppercase\expandafter{\romannumeral2}. In Section \uppercase\expandafter{\romannumeral3}, we provide an overview of the proposed multi-IMU DR system. The preliminaries are described in Section \uppercase\expandafter{\romannumeral4}, including the installation of the multiple IMUs, the coordinate systems definitions, and the error-state model of the EKF. In Section \uppercase\expandafter{\romannumeral5}, we deduce the wheel velocity measurements for different IMUs and the multi-IMU spatial constraint measurement model. Experimental results are explained and analyzed in Section \uppercase\expandafter{\romannumeral6}. Section \uppercase\expandafter{\romannumeral7} summarizes some conclusions and possible directions for future research.

\section{Relevant Work}
\subsection{Wheel-mounted Inertial Sensors}
To provide a low-cost alternative for the wheel odometer, researchers have proposed various approaches to calculate the wheel speed or traveled distance of the vehicle using the wheel-mounted inertial sensors. In \cite{coulter2011}, the authors mounted a tri-axial accelerometer on the wheel to measure the wheel revolutions, rotation angle, and duration of movement of the wheelchairs. Authors in \cite{gersdorf2013} used two accelerometers and a single gyroscope which were mounted on the wheel to estimate the odometry via EKF. The two accelerometers were placed perpendicularly on the wheel plane, while the sensitive axis of the gyroscope was set orthogonal to the wheel plane. The process model of this system was completely driven by the state, thus both the accelerometers and gyroscope were modeled as measurements. An improved single accelerometer-based wheel odometer was proposed in \cite{naser2021sensors}. The peaks and valleys of the accelerometer signal were detected to determine the number of complete cycles covered by the wheel, so as to calculate the traveled distance. Nonetheless, these studies only focused on providing the velocity information instead of the full ego-motion estimation of the wheeled vehicles.

Collin \textit{et al} \cite{collin2014tim, collin2014tvt, mezentsev2019} proposed a 2D DR system using the Wheel-IMU. In this approach, the outputs of the two accelerometers parallel to the wheel plane were used to determine the wheel rotation angle and thereby the traveled distance, while the gyroscope data were leveraged to calculate the vehicle heading. This algorithm had to assume that the vehicle was moving at a constant speed and failed to handle the misalignment error between the Wheel-IMU and wheel center.


\subsection{Multiple IMUs Fusion}
Few studies have been conducted with a focus on integrating multiple IMUs mounted on different places for vehicular navigation. In \cite{naser2020VTC}, the authors mounted two IMUs on the two rear wheels of the land vehicle to provide wheel velocity and vehicle heading change measurements, so as to mitigate the error drift of the onboard IMU during GNSS outages. In the field of IMU array-based localization, a large number of papers have considered the observation-level approaches to fuse the outputs of several IMUs to generate a single virtual IMU measurement \cite{sukkarieh2000multiIMU,zhang2020,bancroft2011}. However, these approaches are not suitable for our problem because, in our design, the IMUs are distributed in different locations on the vehicle to perceive different dynamic information. It is worth mentioning that Bancroft \cite{bancroft2009} designed a centralized filter where the navigation solutions from multiple IMUs (which were placed closely on the vehicle body) were fused together. At the same time, the relative position, velocity, and attitude updates were all introduced to improve the localization accuracy in GNSS-degraded areas.

Another research area about fusing multiple IMUs on multiple rigid bodies is human motion capture. The idea is to estimate the human body kinematics by fusing the data from multiple IMUs mounted on different places of the body, e.g., the shanks, thighs, sacrum, and so on \cite{Fasel2018, Mohamed2020, Vargas2021, lijie2021IOT}. However, these multi-IMU-based human motion capture systems mainly focus on tracking the human pose rather than positioning and navigation. Although a full human motion capture system based on wearable inertial sensor networks which could estimate both the limb motion and trajectory was proposed in \cite{lijie2021IOT}, the sensor fusion algorithm was only adopted to measure the orientations while the localization of the human body was estimated by zero velocity update (ZUPT)-aided INS through EKF. 

The idea to fuse diverse motion information perceived by multiple IMUs and take advantage of the relative position constraints is similar to that encountered in the dual foot-mounted IMUs (Foot-IMUs)-based pedestrian dead reckoning (PDR) system \cite{skog2012,liyu2019,shi2017,laverne2011}. There are two main approaches to implementing the relative restraint between the two IMUs in the dual Foot-IMUs-based PDR: inequality constraint and equality constraint. Based on the fact that there is an upper limit on the distance between the two Foot-IMUs, the authors in \cite{skog2012} proposed a representative inequality constraint algorithm to limit the INS error drift. Niu \textit{et al} \cite{liyu2019} exploited the fact that there is always a minimum distance between the two feet of a person during every gait cycle to formulate the observation model via equality constraint. Bancroft \cite{bancroft2011} compared a centralized filter and a federated filter-based GNSS/multi-IMU integrated navigation system in the context of pedestrian navigation. The centralized filter could provide relatively better results, but required considerably more processing time.    

Nonetheless, unlike the dual Foot-IMUs, the relative positions of the IMUs mounted on the wheels center and vehicle body are fixed in three dimensions. Therefore, one can straightforwardly leverage the three-dimensional position relationship rather than the distance (one dimension) to constrain the multiple IMUs' states. The 3D spacial relationship contains richer information thus would make the constraint tighter and more reliable. In the above-mentioned multi-IMU integrated systems, the IMUs' states are almost all combined into one filtering framework. However, if more states of an IMU are estimated or more IMUs are integrated, the dramatic increase in matrix size would lead to a heavy computational burden \cite{bancroft2011}. For the sake of efficiency and scalability to more IMUs, we adopt a decentralized EKF structure in the proposed multi-IMU system, which is explained in the next section. 

\section{Overview of the Proposed System}
\begin{figure}[htbp]
	\centering
	\subfigure[Dual Wheel-INS]{
		\includegraphics[width=8.5cm]{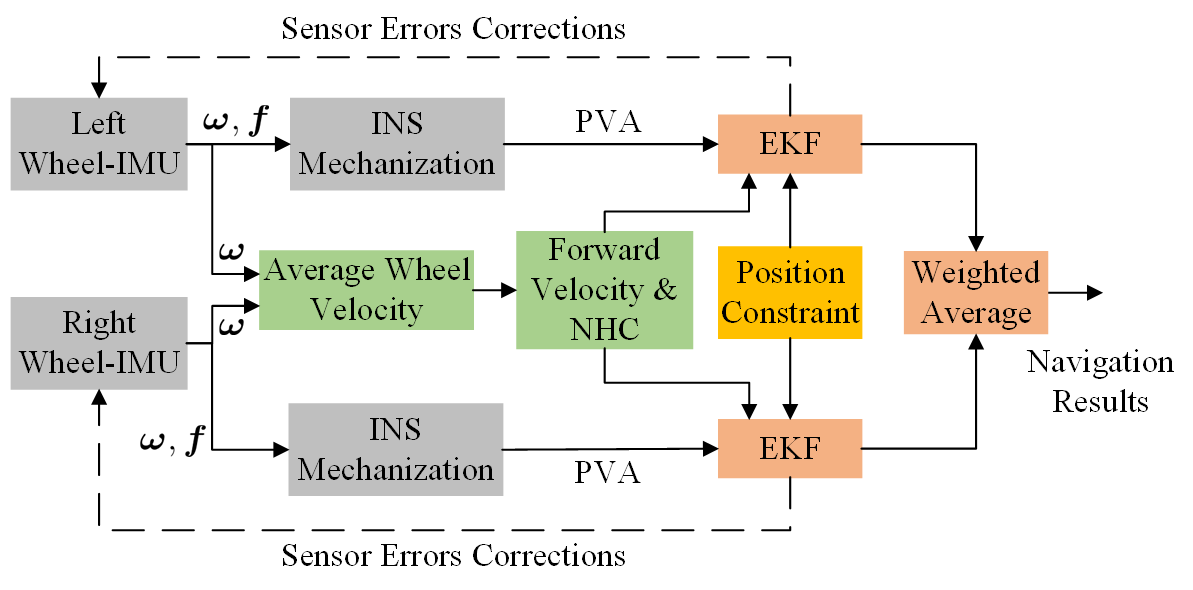}
	}
	\quad
	\subfigure[Body/Wheel-INS]{
		\includegraphics[width=8.5cm]{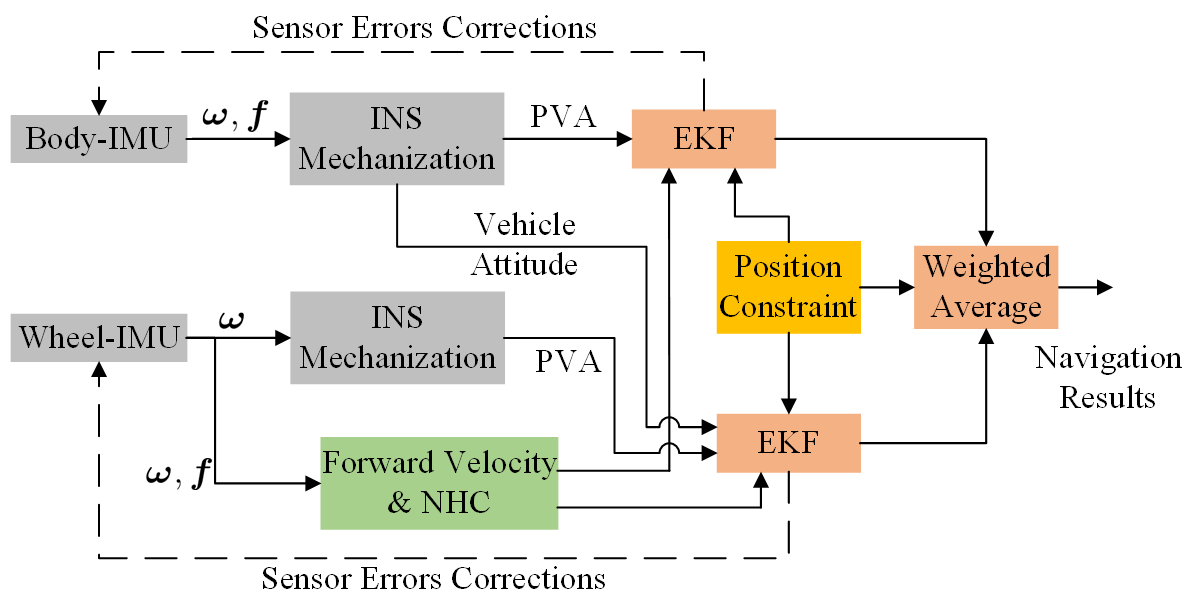}
	}
	\quad
	\subfigure[Triple INS]{
		\includegraphics[width=8.5cm]{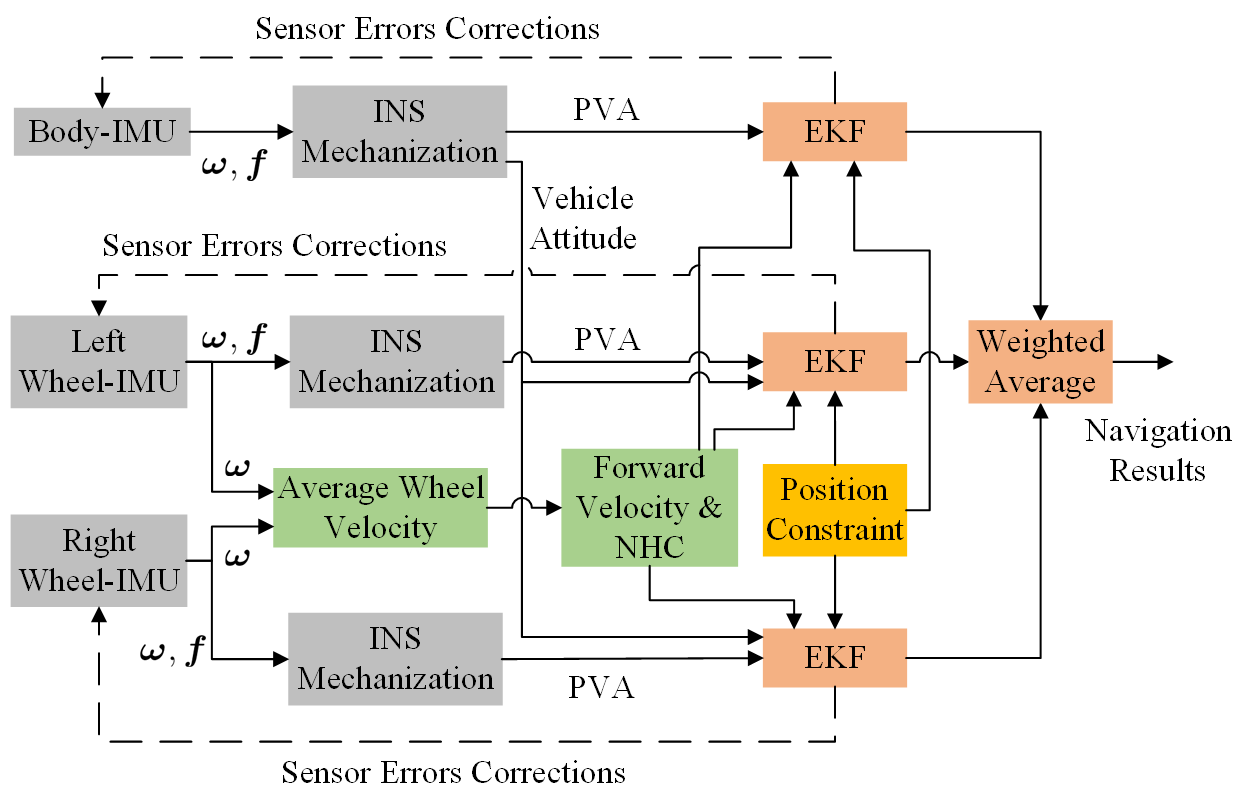}
	}
	\caption{Overview of the algorithm structures of the three different multi-IMU configurations. $\bm\omega$ and $\bm{f}$ are the angular rate and specific force measured by the IMU, respectively; ``PVA'' indicates the position, velocity, and attitude of the IMU.}
	\label{2}
\end{figure}

The key objective of this study is to extend our previous research on single Wheel-IMU based DR system \cite{niu2019} to multi-IMU based DR systems. By mounting multiple MEMS IMUs to different locations of the vehicle, diverse dynamic information is allowed to be exploited and fused to improve positioning accuracy and robustness. Note that at least one IMU has to be mounted on the vehicle wheel to obtain the wheel speed and take advantage of the inherent rotation modulation. For details of Wheel-INS please refer to \cite{niu2019, wu2020arxiv}. We design and implement three typical configurations in this study: Dual Wheel-INS, Body/Wheel-INS, and Triple INS (refer to Table \ref{Tab1}). The algorithm structures of the three DR systems are depicted in Fig. \ref{2}.

The proposed multi-IMU-based DR system is composed of multiple EKF systems corresponding to the states of each IMU. In other words, each IMU maintains its own filter system (including the state model and measurement model) and shares information with each other in this system. For the sake of simplicity, in this paper, the filtering system maintained by the Body-IMU is referred to as ``Body-INS". In each subsystem, the forward INS mechanization is performed to predict the state of each IMU. In the Dual Wheel-INS, the wheel velocities calculated by the gyro outputs of the two Wheel-IMUs and wheel radius are averaged as odometry measurements for each filtering system. In the Body/Wheel-INS, the Wheel-IMU provides wheel velocity for the two subsystems, while the horizontal attitude of the vehicle indicated by the Body-IMU is sent to the Wheel-INS to remove the assumption that the vehicle is moving on the horizontal plane \cite{niu2019}. In the Triple INS, we also adopt the average velocity calculated by the dual Wheel-IMUs. At the same time, the vehicle attitude generated by the Body-INS is shared for the two Wheel-INSs. NHC is integrated with all the subsystems. The spatial constraint between the multiple IMUs is implemented by the coordinates update of a reference point (derived in Section \uppercase\expandafter{\romannumeral5}) to correct each IMU's state, respectively. Finally, the mean heading of the multiple IMUs and the reference point's position are output as the final navigation results. This algorithm structure is somewhat similar to the federated filter \cite{carlson1990, carlson2002}, but there is no master filter in our system. 

We don't adopt the federated Kalman filter for two major reasons. Firstly, in the federated Kalman filter, the local filters contain the common system state; thus, usually, it is the kinematic process noise to be shared \cite{gao1993}.  The federated Kalman filter is proposed to solve this multi-sensor fusion problem in a parallel way without being affected by the cross-correlation issue caused by the common process noise. In \cite{bancroft2011}, the authors first implemented the federated Kalman filter incorporating multiple IMUs which were placed close to each other (on the same rigid body). In this system, the shared states were position, velocity, and attitude. However, in our multi-IMU fusion problem, the IMUs are mounted at different places and thus undergo different motions. The sub-filters share neither process noise nor state directly as they are located at different places in the vehicle and they have unique stochastic errors. Therefore, the federated Kalman filter is not suitable for our case. Secondly, a centralized filter represents the best accuracy one can do in this multi-sensor fusion system \cite{carlson2002}. Therefore, we conducted experiments and compared the proposed distributed filter with the centralized filter \cite{gao1993} (cf. Section VI-B-2). The experimental results illustrated that the proposed distributed filter shows trivial accuracy degradation while exhibiting significant computation efficiency compared to the centralized filter.

\section{Prerequisites}

\subsection{Installation of the Multi-IMU}
\begin{figure}[htbp]
	\centering
	\includegraphics[width=8.6cm]{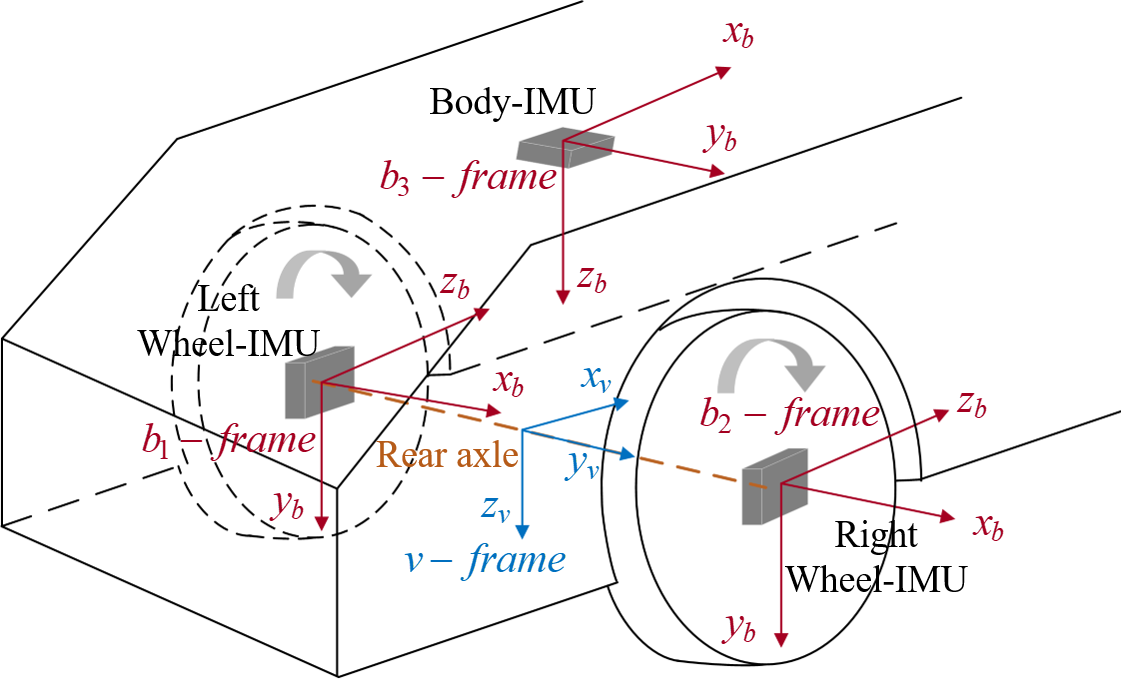}
	\caption{Installation relationship of the multiple IMUs and definitions of the axis directions for the vehicle frame (\textit{v}-frame) and the IMU frame (\textit{b}-frame). All the three IMU frames are defined as \textit{b}-frame here and we use $\{1,2,3\}$ to represent different IMUs as in Fig. \ref{1}.}
	\label{3}
\end{figure}

Fig. \ref{3} illustrates the installation relationship of the multiple IMUs and the definitions of the involved coordinate systems. The Wheel-IMU is mounted on non-steering wheel of the land vehicle so that the vehicle state indicated by Wheel-INS is not affected by the vehicle maneuvers. The definitions of the coordinate frames are listed in Table \ref{Tab2}. 

Fusing the measurements in the \textit{v}-frame (wheel velocity and NHC) with INS requires the knowledge of the installation relationship between the IMU and the host vehicle \cite{chen2020}. However, the misalignment problem is inevitable in any practical applications, which, if not handled properly, could noticeably undermine the system performance \cite{wu2009}. The misalignment error between the IMU and the vehicle can be represented by the lever arm and mounting angles. The lever arm represents the vector from the IMU center to the wheel center projected in the \textit{b}-frame. The mounting angles are the two Euler angles which indicate that the \textit{y}-\textit{z} plane of the \textit{b}-frame of the Wheel-IMU is not parallel to the wheel plane. Refer to \cite{niu2019} for more details on the misalignment errors of the Wheel-IMU. In our experiments, we used the method proposed in \cite{chen2019} to calibrate the Wheel-IMU mounting angles and the lever arm was measured manually. In the following discussion, we assume that all the misalignment errors of both the Wheel-IMU and Body-IMU have been effectively calibrated in advance. Details about the rotation characteristics of the Wheel-IMU and the influence of the misalignment errors are explained in our previous paper \cite{niu2019}.


\begin{table}[h]
	\renewcommand\arraystretch{1.5}
	\centering
	\caption{Definitions of the Coordinates Systems}
	\label{Tab2}
	\begin{tabular}{p{1.6cm}p{3cm}p{3cm}}
		\toprule
		\multicolumn{1}{c}{Symbol} & \multicolumn{1}{c}{Description} & \multicolumn{1}{c}{Definition} \\
		\midrule
		{\textit{v}-frame} & {The vehicle coordinates system.} & 
			{origin: midpoint of the rear axle.\newline
			\textit{x}-axis: longitudinal direction of the host vehicle.\newline
			\textit{y}-axis: right.\newline
			 \textit{z}-axis: down.}
		\\
		{\textit{b}-frame \newline(Wheel-IMU)} & {The coordinate system of the Wheel-IMU.} & {origin: IMU measurement center.\newline
				\textit{x}-axis: right, parallel to the rotation axis.\newline
				 \textit{y}- and \textit{z}-axes are parallel to the wheel surface to complete a right-handed orthogonal frame.}
		\\
		{\textit{b}-frame \newline(Body-IMU)} & {The coordinate system of the Body-IMU.} & {origin: IMU measurement center.\newline
			\textit{x}-axis: forward direction.\newline
			\textit{y}-axis: right.\newline
			\textit{z}-axis: down.}
		\\
		{\textit{n}-frame} & {The local level frame.} & {origin: the same as \textit{b}-frame.\newline
			\textit{x}-axis: north.\newline
			\textit{y}-axis: east.\newline
			\textit{z}-axis: downward vertically.}
		\\
		{\textit{w}-frame} & {The world frame for the local DR system.} & {Aligned with the initial \textit{n}-frame determined by INS. Each IMU maintains its own \textit{w}-frame.}
		\\
		\bottomrule
	\end{tabular}
\end{table}

\subsection{Error State Model}
For the sake of computation efficiency, we do not integrate the states from the multiple IMUs into one state vector and maintain only one filter system. The proposed multiple IMUs-based DR system consists of multiple Kalman filters which are built in a distributed structure. Each IMU retains its own state. They share information, e.g., wheel speed, vehicle attitude, and IMU positions with each other and update their own state independently.   

In this study, we choose the 21-dimensional error-state model for every IMU. It was illustrated in \cite{niu2019} that the 21-state could achieve better performance. The conventional strapdown inertial navigation is leveraged to predict the IMU state. The kinematic equations for the strapdown INS have been described extensively in the literature \cite{britting1971,Shin2005,Yan2019}. In this study, the error-state vector of the EKF is constructed in the \textit{n}-frame, including three dimensional position errors, three dimensional velocity errors, attitude errors, the residual bias and scale factor errors of the gyroscope and accelerometer, which can be written as
\begin{equation}
\bm{x}(t)=\left[\left(\delta \bm{r}^{n}\right)^{\mathrm{T}} \quad\left(\delta \bm{v}^{n}\right)^{\mathrm{T}} \quad \bm{\phi}^{\mathrm{T}} \quad \bm{b}_{g}^{\mathrm{T}} \quad \bm{b}_{a}^{\mathrm{T}} \quad \bm{s}_{g}^{\mathrm{T}} \quad \bm{s}_{a}^{\mathrm{T}} \right]^{\mathrm{T}}
\end{equation}
where $\delta \bm{r}^{n}$, $\delta \bm{v}^{n}$ and $\bm{\phi}$ indicate the position, velocity and attitude errors of INS, respectively; $\bm{b}_{g}$ and $\bm{b}_{a}$ denote the residual bias errors of the gyroscope and the accelerometer, respectively; $\bm{s}_{g}$ and $\bm{s}_{a}$ are the residual scale factor errors of the gyroscope and accelerometer, respectively. We adopt a simplified phi-angle error model of INS as the kinematic model. The first-order Gauss-Markov model \cite{maybeck1982stochastic} is employed to establish the state model of the residual sensor errors. In summary, the state model employed in the error-state EKF can be written as,
\begin{equation}
\left\{
\begin{aligned}
\delta \dot{\bm{r}}^{n}&=\delta \bm{v}^{n} \\
\delta \dot{\bm{v}}^{n}&=-\bm{\phi} \times \textbf{C}_b^n\bm{f}^{b} + \textbf{C}_{b}^{n}\delta\bm{f}^{b} \\
\dot{\bm{\phi}}&=-\textbf{C}_{b}^{n}\delta\bm{\omega}_{ib}^{b}\\
\dot{\bm{b}}_{g}&=-(1/\tau_{bg}) \bm{b}_{g}+\bm{w}_{bg}\\
\dot{\bm{b}}_{a}&=-(1/\tau_{ba}) \bm{b}_{a}+\bm{w}_{ba}\\
\dot{\bm{s}}_{g}&=-(1/\tau_{sg}) \bm{s}_{g}+\bm{w}_{sg}\\
\dot{\bm{s}}_{a}&=-(1/\tau_{sa}) \bm{s}_{a}+\bm{w}_{sa}
\end{aligned}
\right.
\end{equation}
where $\textbf{C}_{b}^{n}$ is the attitude matrix of the IMU with respect to (w.r.t.) the \textit{n}-frame; $\delta\bm{\omega}_{ib}^{b}$ and $\delta\bm{f}^{b}$ are the errors of the gyroscope and accelerometer measurements, respectively, which can be expressed as $\delta\bm{\omega}_{ib}^{b}= \bm{b}_{g}+\mathrm{diag}(\bm{\omega}_{ib}^{b})\bm{s}_{g}+\bm{n}_g$ and $\delta\bm{f}^{b}= \bm{b}_{a}+\mathrm{diag}(\bm{f}^{b})\bm{s}_{a}+\bm{n}_a$, respectively, where $\bm{n}_g$ and $\bm{n}_a$ are the measurement white noise of the gyroscope and accelerometer; $\mathrm{diag}(*)$ is the diagonal matrix form of a vector; $\times$ indicates the cross product of two vectors. Finally, $\bm{w}_{bg}$ and $\bm{w}_{ba}$ denote the driving white noise of the residual bias errors of the gyroscope and accelerometer, respectively; $\bm{w}_{sg}$ and $\bm{w}_{sa}$ denote the driving white noise of the scale factor errors of the gyroscope and accelerometer, respectively; $\tau_{bg}$, $\tau_{ba}$, $\tau_{sg}$, and $\tau_{sa}$ are the correlation time in the first-order Gauss-Markov model of the gyroscope bias, gyroscope scale factor, accelerometer bias, and accelerometer scale factor, respectively. The continuous-time dynamic model and Jacobian matrix of the EKF can be found in \cite{niu2019}. This error-state model was employed for all the subsystems. Because of the limitation of space, the initial error states covariance and process noise covariance are not given in detail here. We encourage the readers to find them in our open-sourced code and reproduce the results.

\section{Measurement Models}
In this section, the derivations of the measurement models in the proposed multiple IMUs-based DR system are described. Firstly, we deduce the 3D wheel velocity observation models for Wheel-INS and Body-INS, respectively. After that, the relative position constraint among the multiple IMUs is formulated and analyzed by taking the Triple INS as instance. 
\subsection{Wheel Velocity Measurement}
In the proposed multi-IMU based DR system, the wheel velocity and NHC are integrated as a 3D velocity measurement in the \textit{v}-frame to update the state of each IMU. By mounting the IMU to the Wheel center, the wheel velocity can be calculated by the wheel radius and the gyro data of the Wheel-IMU, which can be expressed as  
\begin{equation}
\begin{aligned}
\widetilde{v}^{v}_{wheel} &=\widetilde{\omega}_{x}r-e_v = ({\omega}_{x}+\delta{\omega}_{x})r-e_v\\ &={v}^{v}_{wheel}+r\delta{\omega}_x - e_v
\end{aligned}
\end{equation}
where $\widetilde{v}^{v}_{wheel}$ and ${v}^{v}_{wheel}$ are the measured and true vehicle forward velocity, respectively; $\widetilde{\omega}_{x}$ is the gyroscope output in the \textit{x}-axis; ${\omega}_{x}$ is the true value of the angular rate in the \textit{x}-axis of Wheel-IMU; $\delta{\omega}_{x}$ is the gyroscope measurement error; $r$ is the wheel radius, and $e_v$ is the forward velocity observation noise, modeled as Gaussian white noise. Integrating with the NHC, the 3D velocity measurement model in the \textit{v}-frame can be written as 

\begin{equation}
\widetilde{\bm{v}}^{v}_{wheel} = \left[ {v}^{v}_{wheel}+r\delta{\omega}_x \quad 0 \quad 0 \right]^{\mathrm{T}} - \bm{e}_v
\end{equation}
where $\bm{e}_v$ is the velocity measurement noise vector, including both the forward velocity noise and NHC noise. The standard deviation of the velocity measurement noise set in the experiments are listed and discussed in Appendix \ref{EstimatorParameters}. As discussed in \cite{niu2019}, because the Wheel-IMU rotates with the wheel, the pitch angle  of the vehicle cannot be indicated by Wheel-INS; thus we have to assume that the vehicle is moving on the horizontal surface in Wheel-INS and Dual Wheel-INS. However, with the integration of Body-IMU, this limitation can be lifted in Body/Wheel-INS and Triple INS. 

Because the state errors contained in the measurement models of Body-INS and Wheel-INS are different, the error perturbations in the two systems also differ \cite{Yan2019}. For example, the attitude of the Body-IMU w.r.t. the vehicle is known and fixed after calibration, while in the Wheel-INS, the IMU heading error is included in the vehicle's attitude. As per Fig. \ref{1}, we use the subscript ``1" and ``3" to represent the Wheel-IMU and Body-IMU, respectively. In consequence, the vehicle velocity calculated by the Wheel-IMU and the Body-IMU can be written as

\begin{equation}
\begin{aligned}
\hat{\bm{v}}^v_{wheel1} &= \hat{\textbf{C}}^v_n \hat{\bm{v}}^n_{b_1} + \hat{\textbf{C}}^v_n \hat{\textbf{C}}^n_{b_1} \left( \hat{\bm{\omega}}^{b_1}_{nb} \times \right) \bm{l}^{b_1}_{wheel}\\
&\approx {\textbf{C}}^v_n (\textbf{I} \!+\! \delta{\bm{\psi}_1} \times)({\bm{v}}^n_{b_1} \!+\! \delta{\bm{v}}^n_{b_1})\\
&\quad +\! {\textbf{C}}^v_n(\textbf{I} \!+\! \delta\bm{\psi}_1 \times)(\textbf{I} - \bm{\phi}_1 \times) \textbf{C}^n_{b_1} ({\bm{\omega}}^{b_1}_{nb}\!\times \!+\! \delta{\bm{\omega}}^{b_1}_{nb} \times) \bm{l}^{b_1}_{wheel}\\
&\approx  {\bm{v}}^v_{wheel1} + \textbf{C}^v_n \delta{\bm{v}}^n_{b_1} -  \textbf{C}^v_n \textbf{C}^n_{b_1} \left( \bm{l}^{b_1}_{wheel} \times \right)\delta\bm{\omega}^{b_1}_{ib}\\
&\quad- \textbf{C}^v_n \left[\left( {\bm{v}}^n_{b_1} \times \right) + \left({\textbf{C}}^n_{b_1} \left( \bm{\omega}^{b_1}_{ib} \times \right) \bm{l}^{b_1}_{wheel} \right) \times \right]\delta\bm{\psi}_1 \\
&\quad+ \textbf{C}^v_n \left[ \left(\textbf{C}^n_{b_1} \left( \bm{\omega}^{b_1}_{ib} \times \right) \bm{l}^{b_1}_{wheel} \right) \times \right]\bm{\phi}_{1}
\end{aligned}
\label{velmeasurement}
\end{equation} 

\begin{equation}
\begin{aligned}
\hat{\bm{v}}^v_{wheel3} &= {\textbf{C}}^v_{b_3} \hat{\textbf{C}}^{b_3}_n \hat{\bm{v}}^n_{b_3} + {\textbf{C}}^{v}_{b_3} \left( \hat{\bm{\omega}}^{b_3}_{nb} \times \right) \bm{l}^{b_3}_{wheel}\\
&\approx \textbf{C}^v_{b_3} {\textbf{C}}^{b_3}_n (\textbf{I} \!+\! \bm{\phi}_3 \times)(\bm{v}^n_{b_3} + \delta{\bm{v}}^n_{b_3}) \\
&\quad + \textbf{C}^v_{b_3} ({\bm{\omega}}^{b_3}_{nb}\!\times \!+\! \delta{\bm{\omega}}^{b_3}_{nb} \times) \bm{l}^{b_3}_{wheel}\\
&\approx {\bm{v}}^v_{wheel3} + \textbf{C}^v_{b_3} {\textbf{C}}^{b_3}_n \delta{\bm{v}}^n_{b_3} - \textbf{C}^v_{b_3} {\textbf{C}}^{b_3}_n \left( {\bm{v}}^n_{b_3} \times \right)\bm{\phi}_{3} \\
&\quad - \textbf{C}^v_{b_3} \left( \bm{l}^{b_3}_{wheel} \times \right)\delta{\bm{\omega}}^{b_3}_{ib}\\
\end{aligned}
\end{equation} 
where the subscripts $b_1$ and $b_3$ indicate the body frame of the Wheel-IMU and Body-IMU, respectively; ${\textbf{C}}^v_{b_3}$ is the attitude misalignment error between the Body-IMU and the vehicle, which can be calibrated using the method proposed in \cite{chen2019} by continuously rotating the wheel around the rotation axis; $\bm{l}^{b}_{wheel}$ is the lever arm between the IMU and the wheel center (refer to \cite{niu2019} for details), which can be either measured manually by a tapeline or estimated online by augmenting it into the state vector; $\delta\bm{\psi}$ is the attitude error vector only including the heading error, i.e., $\delta\bm{\psi} = \left[ 0 \!\quad \!0 \! \quad \! \delta{\psi} \right]^{\mathrm{T}}$; ${\textbf{C}}^v_n$ is the vehicle attitude matrix, which can be transformed from the vehicle Euler angles $\bm{\vartheta}_v^n$. In Dual Wheel-INS, because of the horizontal motion assumption of the vehicle, we have ${\bm{\vartheta}}_v^n = \left[ 0 \!\quad \!0\! \quad \!\phi_{1}\!-\!\pi/2 \right]^{\mathrm{T}}$, namely, the pitch and roll angles of the vehicle are assumed to be zero; whereas in the Body/Wheel-INS, with the horizontal angles of the vehicle indicated by Body-INS, we have  ${\bm{\vartheta}}_v^n = \left[ \phi_{3} \!\quad \!\theta_{3} \! \quad \!\phi_{1}\!-\!\pi/2 \right]^{\mathrm{T}}$, where $\phi_3$ and $\theta_3$ are the roll and pitch angles of the vehicle indicated by Body-INS, and $\psi_1$ is the heading of the vehicle indicated by Wheel-INS. The final velocity error measurement equations of Wheel-INS and Body-INS can be obtained by subtracting Eq. 4 from Eq. 5 and Eq. 6. The design matrix of the velocity measurement in left Wheel-INS is given in Appendix \ref{DesignMatrices}.

\subsection{Multi-IMU Position Constraint}
\begin{figure}[htbp]
	\centering
	\includegraphics[width=8.6cm]{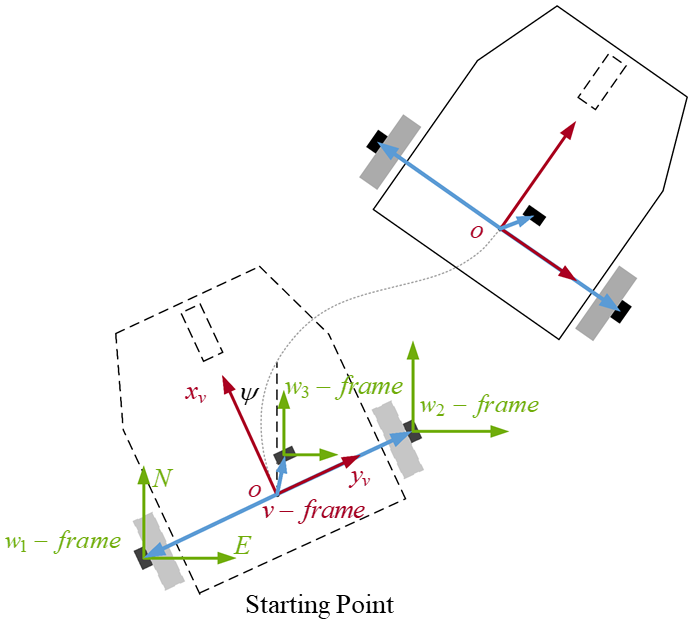}
	\caption{Illustration of the position constraint among the multiple IMUs (top view). The multiple IMUs' positions in the \textit{v}-frame (the blue arrows) keep constant during the movement of the vehicle.${\textit{w}_1}$-frame, ${\textit{w}_2}$-frame, and ${\textit{w}_3}$-frame indicate the world frame determined by the left Wheel-IMU, right Wheel-IMU, and Body-IMU, respectively.}
	\label{4}
\end{figure}

Given that the multiple IMUs are rigidly installed, their relative positions remain fixed during the movement of the vehicle. This condition can be leveraged as a measurement update for the filter systems. Note that this constraint can be equivalent to that the coordinates of the multiple IMUs in the \textit{v}-frame are constant. Unlike the equality and inequality constraint in the dual Foot-IMUs-based PDR, the position constraint in the proposed multi-IMU-based DR system is in three dimensions. To keep the computational complexity at a relatively low level, we adopt a distributed EKF structure rather than accumulating the states from every single IMU into one state vector to form one filtering system. Each filter system works independently. The 3D spatial constraint between the multiple IMUs is treated as an external observation for all the subsystems. Fig. \ref{4} illustrates the relative position constraints between the multiple IMUs. Here, we use the left Wheel-IMU in the Triple INS (dual Wheel-IMUs plus one Body-IMU) as an example to show the deviation of the measurement model of the 3D relative position constraint.

\begin{figure}[!t]
	\label{alg1}
	\renewcommand{\algorithmicrequire}{\textbf{Input:}}
	\renewcommand{\algorithmicensure}{\textbf{Output:}}
	\removelatexerror
	\begin{algorithm}[H]
		\caption{Multi-IMU Position Constraint Measurement}
		\begin{algorithmic}[1]
			\REQUIRE Multiple IMUs' positions in the \textit{w}-frames indicated by the subsystems, i.e., $\hat{\bm{r}}_{b_1}^{w_1}$, $\hat{\bm{r}}_{b_2}^{w_2}$, and $\hat{\bm{r}}_{b_3}^{w_3}$
			\ENSURE Reference point's position in the \textit{w}-frames, i.e., $\hat{\bm{r}}_{o}^{w_1}$, $\hat{\bm{r}}_{o}^{w_2}$, and $\hat{\bm{r}}_{o}^{w_3}$
			\STATE Calculate the coordinates of the reference point $o$ in the \textit{w}-frames maintained by different IMUs (cf. Eq. \ref{oinwframe}).

			\STATE Transform the coordinates of the reference point $o$ in the other \textit{w}-frames to the objective \textit{w}-frame (cf. Eq. \ref{weightedpos} and Eq. \ref{wframeTrans}).

			\STATE Return the weighted average position of the reference point in the objective \textit{w}-frame (cf. Eq. \ref{weightedpos}).
		\end{algorithmic}
	\end{algorithm}
\end{figure}

As the DR results of the subsystems are represented in the \textit{w}-frame, the relative position constraint among the multiple IMUs is also constructed in the \textit{w}-frame. According to the definition of the \textit{w}-frame (cf. Table \ref{Tab1}), it is known that each IMU maintains its \textit{w}-frame and there is only position difference between the \textit{w}-frames determined by the multiple IMUs, shown as Fig. \ref{4}. The translations between the \textit{w}-frames can be calculated by the vehicle's initial heading and the positions of the IMUs in the \textit{v}-frame (referring to Eq. \ref{wframeTrans}). The position constraint measurement is established by determining the coordinate of the reference point (which is treated as external position observation) by the subsystems together. An overview of the algorithm is given in Algorithm 1. 

We choose the origin of the \textit{v}-frame $o$, i.e., the midpoint of the rear axle, as the reference point. Afterward, the coordinate of the reference point $o$ in each \textit{w}-frame can be calculated. Finally, the weighted average of the coordinates of the reference point $o$ in each \textit{w}-frame (referring to Eq. \ref{weightedpos}) is calculated as the external position observation to correct the states of the subsystems. This is similar to the GNSS coordinate update in the GNSS/INS integrated system \cite{Shin2005}. However, in this case, the reference position is generated by multiple IMUs rather than an external sensor. The position of the reference point $o$ in the \textit{w}-frame calculated by the multiple IMUs can be written as 
\begin{equation}
\hat{\bm{r}}_o^{w_i} = \hat{\bm{r}}_{b_i}^{w_i} - \hat{\textbf{C}}_v^{w_i} {\bm{r}}^v_{b_i}
\label{oinwframe}
\end{equation}
where $i\in\{1,2,3\}$ which denotes the IMU number; $w_i$ indicates the \textit{w}-frame established by th \textit{i}-th IMU; ${\bm{r}}^w_{o}$ is the position of the reference point $o$ in the \textit{w}-frame; ${\bm{r}}^w_{b}$ is the position of the IMU in the corresponding \textit{w}-frame; $\hat{\textbf{C}}_v^{w}$ is the attitude matrix of the vehicle w.r.t. the \textit{w}-frame, which is the same as $\hat{\textbf{C}}_v^{n}$; ${\bm{r}}^v_{b}$ is the IMU's position in the \textit{v}-frame, which can be measured in advance. Here we take the left Wheel-IMU as example. The weighted average position of the reference point $o$ in the \textit{w}-frame maintained by the left Wheel-IMU can then be written as
\begin{equation}
\widetilde{\bm{r}}_o^{w_1} =  \bm{k}_{1} {\bm{r}}_{o}^{w_1} + \bm{k}_{2} \left( {\bm{r}}_{o}^{w_2} + \bm{T}^{w_1}_{w_2}\right) + \bm{k}_{3} \left( {\bm{r}}_{o}^{w_3} + \bm{T}^{w_1}_{w_3}\right)
\label{weightedpos}
\end{equation}
where $\bm{k}_{1}$, $\bm{k}_{2}$, and $\bm{k}_{3}$ are the weights of the three systems; $\bm{T}^{w_1}_{w_2}$ is the translation vector between the two \textit{w}-frames maintained by the two Wheel-IMUs, which can be calculated by
\begin{equation}
\bm{T}^{w_1}_{w_2} = \!\begin{bmatrix}
\mathrm{cos}{\psi} &\! \!\mathrm{sin}{\psi} &\!0\!\\ 
-\mathrm{sin}{\psi} &\! \!\mathrm{cos}{\psi} &\!0\!\\ 
0 &\! \!0 &\!1\!
\end{bmatrix} \left( {\bm{r}}^v_{b_1} - {\bm{r}}^v_{b_2} \right)
\label{wframeTrans}
\end{equation}
where $\psi$ is the initial heading of the vehicle. $\bm{T}^{w_1}_{w_3}$ is the translation vector between the two \textit{w}-frames maintained by the left Wheel-IMU and Body-IMU, which can be calculated in the similar way as Eq. \ref{wframeTrans}. Additionally, the weighted position of the reference point $o$ in the \textit{w}-frame maintained by other IMUs can be derived in a similar way. In Dual Wheel-INS or Body/Wheel-INS, the positions of the reference point are calculated by the two IMUs mounted on the vehicle. We roughly model the measurement noise of the reference point position $\bm{e}_{mlt}$ as the Gaussian white noise; thus, we have
\begin{equation}
\widetilde{\bm{r}}_o^{w_1} = {\bm{r}}_o^{w_1}- \bm{e}_{mlt}
\end{equation}

Performing the perturbation analysis of Eq. \ref{oinwframe}, the observation equation of the multi-IMU relative position constraint in the left Wheel-INS can be written as
\begin{equation}
\begin{aligned}
\delta{\bm{z}_{mlt}} &= \hat{\bm{r}}_o^{w_1} \!-\! \widetilde{\bm{r}}_o^{w_1}\\
&={\bm{r}}_{b_1}^{w_1} \!+\! \delta{\bm{r}}_{1}^{n} \!-\! \left[ \textbf{I} \!-\! \left( {\bm{\phi}}_1 \times \right)\right]\textbf{C}_v^{w_1}{\bm{r}}_{b_1}^{v} \!-\! {\bm{r}}_{o}^{w_1} \!+\! \bm{e}_{mlt}\\
&=\delta{\bm{r}}_{1}^{n} \!-\! \left( \textbf{C}_v^{w_1}{\bm{r}}_{b_1}^{v} \right) \times {\bm{\phi}}_1 \!+\! \bm{e}_{mlt}
\end{aligned}
\label{mltIMUmeasurement}
\end{equation}
where $\textbf{I}$ is the identity matrix. The observation equations of the other two IMUs can be obtained in the same way. The design matrix of the multi-IMU position constraint measurement in left Wheel-INS is given in Appendix \ref{DesignMatrices}.

In the proposed multi-IMU-based DR algorithm, each IMU retains its own filtering system and exchanges information mutually with a constant frequency. This design aims to improve the robustness and accuracy of the DR system against the single Wheel-INS while maintaining the computational cost at a reasonable level. However, according to the derivation of the multi-IMU position constraint measurement model above, it is known that the observation information is generated by the system states; thus, the measurement and the state in the EKF are correlated. Additionally, it is rough to simply regard the observation noise as Gaussian white noise which would result in overly optimistic standard deviations (STD) of the estimates. As the uncertainty of the subsystems all tends to be optimistic, we cannot guarantee that their ratios can reflect the real situation. Hence, it is not reasonable to use the calculated STD of each estimated IMU position as the weight to determine the reference point's position (cf. Eq. \ref{weightedpos}). One solution to this issue is to refer to the error level of the inertial sensor, for example, setting the weight of each system according to the gyro bias of the corresponding IMU. Considering that the IMUs used in our experiments were at the same level (cf. Section \uppercase\expandafter{\romannumeral6}), we let the weights of the subsystems be equal, namely, $\bm{k}_{1}\!=\!\bm{k}_{2}\!=\!\bm{k}_{3}\!=\!1/3 \textbf{I}$ in Triple INS. This is statistically valid. Note that the weights can also be determined based on experimental experience. 

Although this solution may be suboptimal, it exhibits excellent scalability. The experimental results also show that it provides navigation results that are competitive with that of the single Wheel-INS (cf. Section \uppercase\expandafter{\romannumeral6}). Moreover, it is noteworthy that the Wheel-IMU should be placed as close as possible to the wheel center to reduce the errors in the multi-IMU position constraint measurement. This is because in the whole wheel plane, only the center point's position is constant in the \textit{v}-frame due to the rotation of the wheel. 


\section{Experimental Results}
This section provides and discusses the experimental results to illustrate the positioning performance of the proposed multi-IMU integrated DR system. Firstly, the experimental conditions, including the experimental platform, test environment, and data processing flow are described. Then, we evaluate the performance of the three IMU configurations (Dual Wheel-INS, Body/Wheel-INS, and Triple INS) using multiple sets of experiments and compared them with the single Wheel-INS. In addition, the comparison between the proposed distributed filter system and the centralized filter in Triple INS is also shown.

\subsection{Experimental Description}

\begin{figure}[htbp]
	\centering
	\includegraphics[width=8.6cm]{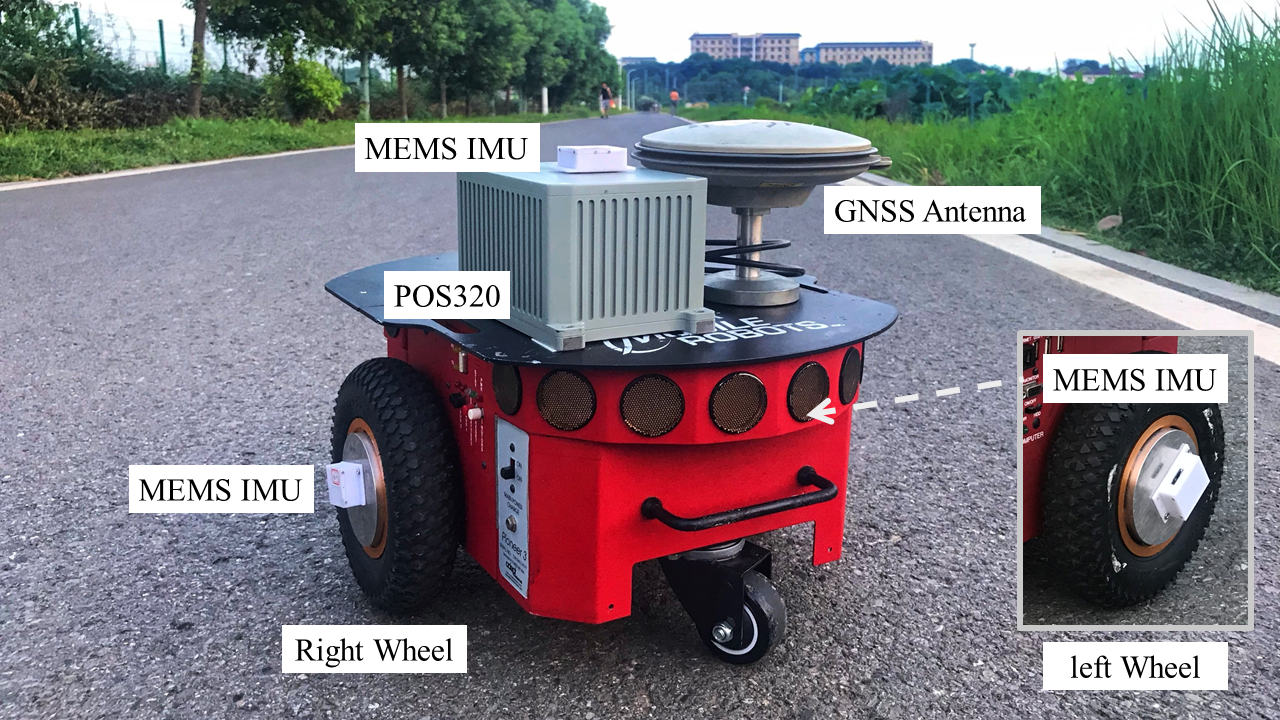}
	\caption{Test equipment and platform.}
	\label{5}
\end{figure}

\begin{figure}[htbp]
	\centering
	\includegraphics[width=8.6cm]{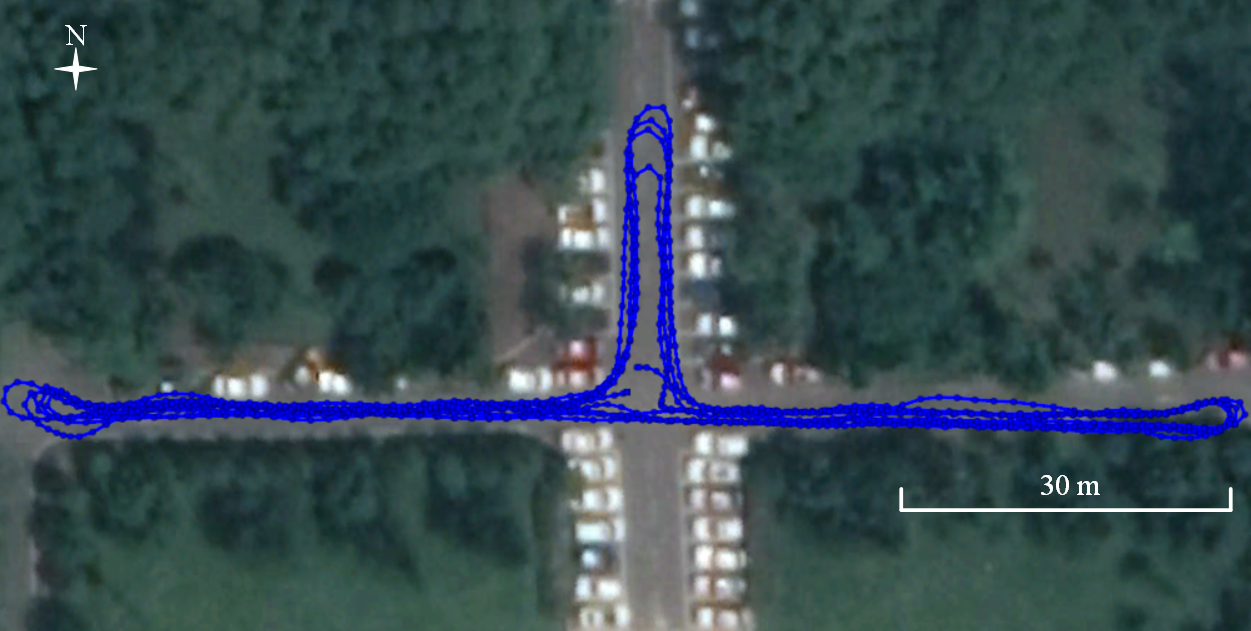}
	\caption{Test trajectory in the campus of Information Department in Wuhan University.}
	\label{6}
\end{figure}

Field tests were conducted on the campus of Wuhan University, Wuhan, China, using a classic differential-drive wheeled robot (Pioneer 3DX). The MEMS IMU modules used in our experiments were customized, containing four ICM20602 (Invensense, TDK Co., Japan) inertial sensor chips which were numbered as C1, C2, C3, and C4. We assume their measurement centers are coincident because the position misalignment errors between them are negligible compared with the relative positions of the IMU modules. The MEMS IMU also included a chargeable battery module for power supply and a Bluetooth module for time synchronization. The multiple MEMS IMUs could be connected with an android mobile phone via Bluetooth to start and end the data collection. Therefore, the outputs of the multiple MEMS IMUs were aligned under one time frame. In our experiments, three MEMS IMUs were carefully placed on the vehicle body and the left and right wheels of the robot, respectively, as shown in Fig. \ref{5}. The robot was also equipped with a high-precision position and orientation system (POS320, MAP Space Time Navigation Technology Co., Ltd., China) which contains a tactical-grade IMU and a GNSS receiver (Trimble Inc., U.S.) to provide the pose ground truth. The main technical parameters of the reference IMU and MEMS IMU are listed in Table \ref{Tab3}. Fig. \ref{5} shows the experimental devices and platform. In the experiments, the robot moved approximately five times along a loop closure trajectory, as shown in Fig. \ref{6}. The total distance traveled by the robot was approximately 1227 m, and the speed of the robot was around 1.4 m/s on average.

\begin{table}[h]
	\centering
	\caption{Technical Parameters of the IMUs Used in the Tests}
	\label{Tab3}
	\begin{threeparttable}
		\begin{tabular}{p{3cm}<{\centering}p{1.5cm}<{\centering}p{1.5cm}<{\centering}}
			\toprule
			IMU & POS320 & ICM20602\\
			\midrule
			Gyro Bias ($deg/h$) & 0.5 & 200\\
			{ARW\tnote{*}  ($deg/\sqrt{h}$)} & 0.05 & 0.24\\
			{Acc.\tnote{*} Bias  ($m/s^2$)} & 0.00025 & 0.01\\
			{VRW\tnote{*}  ($m/s/\sqrt{h}$)} & 0.1 & 3\\
			{Gyro scale factor std\tnote{*}} & 0.001 & 0.03\\
			{Acc. scale factor std} & 0.001 & 0.03\\
			\bottomrule
		\end{tabular}
		\begin{tablenotes}   
			\footnotesize            
			\item[*]ARW denotes the angle random walk; Acc. denotes accelerometer; VRW denotes velocity random walk; std denotes standard deviation.      
		\end{tablenotes}            
	\end{threeparttable}
\end{table}

We collected the data from all four chips inside one MEMS IMU module as four sets of test data for postprocessing. The reference data from the high-end IMU and GNSS receiver were processed through a smoothed post-processed kinematic (PPK)/INS integration method with the data from a nearby reference station. Before data processing, the attitude misalignments of both the Body-IMU and Wheel-IMU were calibrated. The lever arms were also measured. Because we mainly focused on the DR performance of the proposed system, the initial headings of the test systems were set by the heading of the reference system directly. However, this is not reasonable in a real robot navigation application, in which other online initial alignment methods should be investigated. The static IMU data at the starting point was used to calculate the initial roll and pitch angles of the IMU as well as the initial value of the gyro bias. The positions of the multiple MEMS IMUs in the \textit{v}-frame were measured manually three times to get the mean values. The velocity and NHC update frequency was set to 2 Hz, and the relative position constraint frequency was set to 1 Hz.

The comparison between Wheel-INS and ODO/INS and the analysis of different measurement models in Wheel-INS have been conducted in our previous studies \cite{niu2019, wu2020arxiv}. Experimental results have illustrated that Wheel-INS outperforms ODO/INS in both terms of accuracy and robustness. Therefore, the experimental analysis in this paper focuses on evaluating the performance of the three different IMU configurations and the comparison between the proposed distributed filter and the centralized filter. 

\subsection{Performance Comparison}

\subsubsection{Performance Comparison among the Three Configurations}
As the outputs of all four inertial chips in one MEMS IMU module were collected, we had eight sets of Wheel-IMU (left and right) data to employ the single Wheel-INS for comparison. In Dual Wheel-INS and Body/Wheel-INS, the data from the two MEMS IMUs were permuted and combined to generate sixteen sets of results.

\begin{figure}[htbp]
	\centering
	\includegraphics[width=8.6cm]{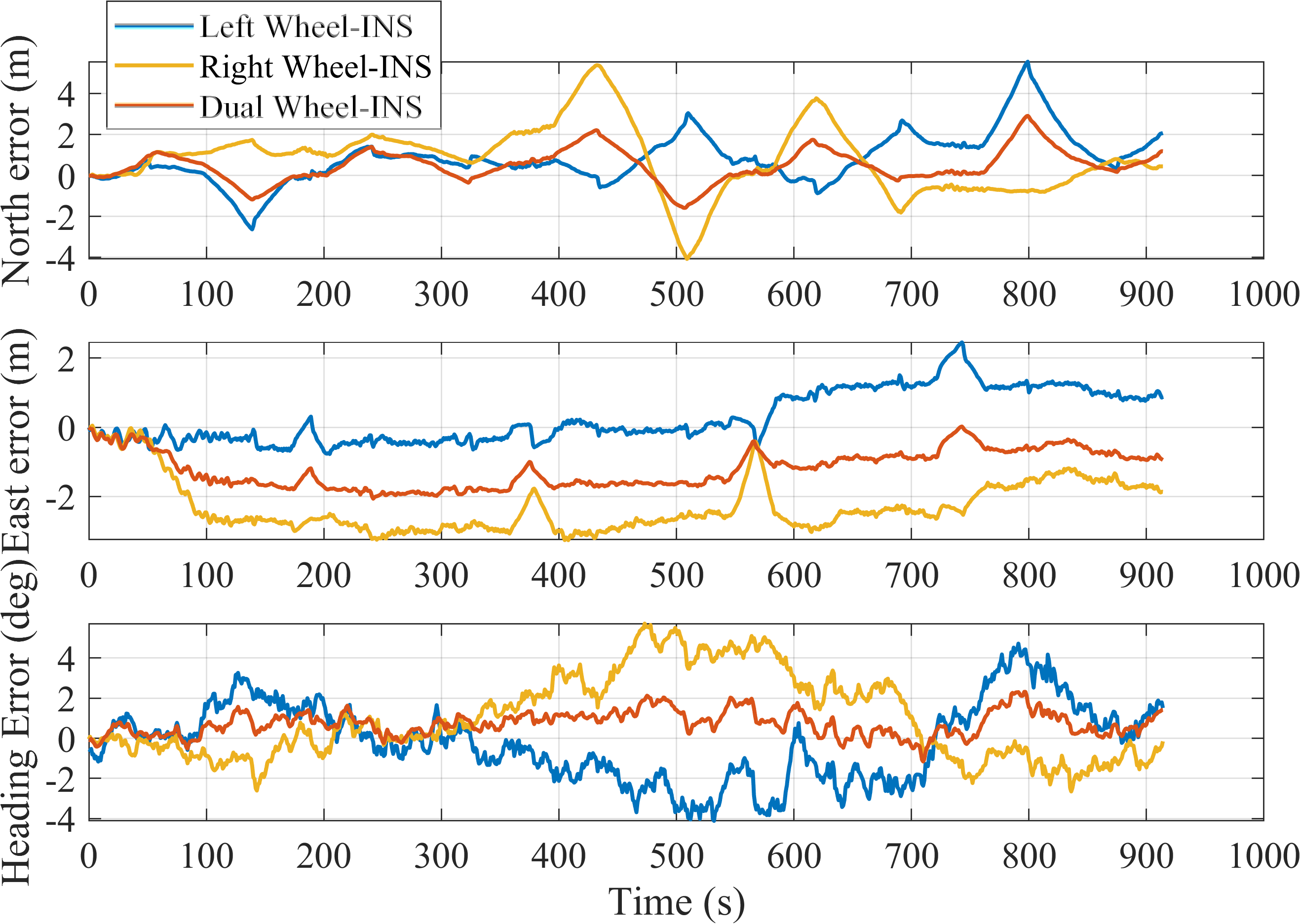}
	\caption{The horizontal positioning error and heading error of Dual Wheel-INS and the two subsystems.}
	\label{7}
\end{figure}

\begin{figure}[htbp]
	\centering
	\includegraphics[width=8.6cm]{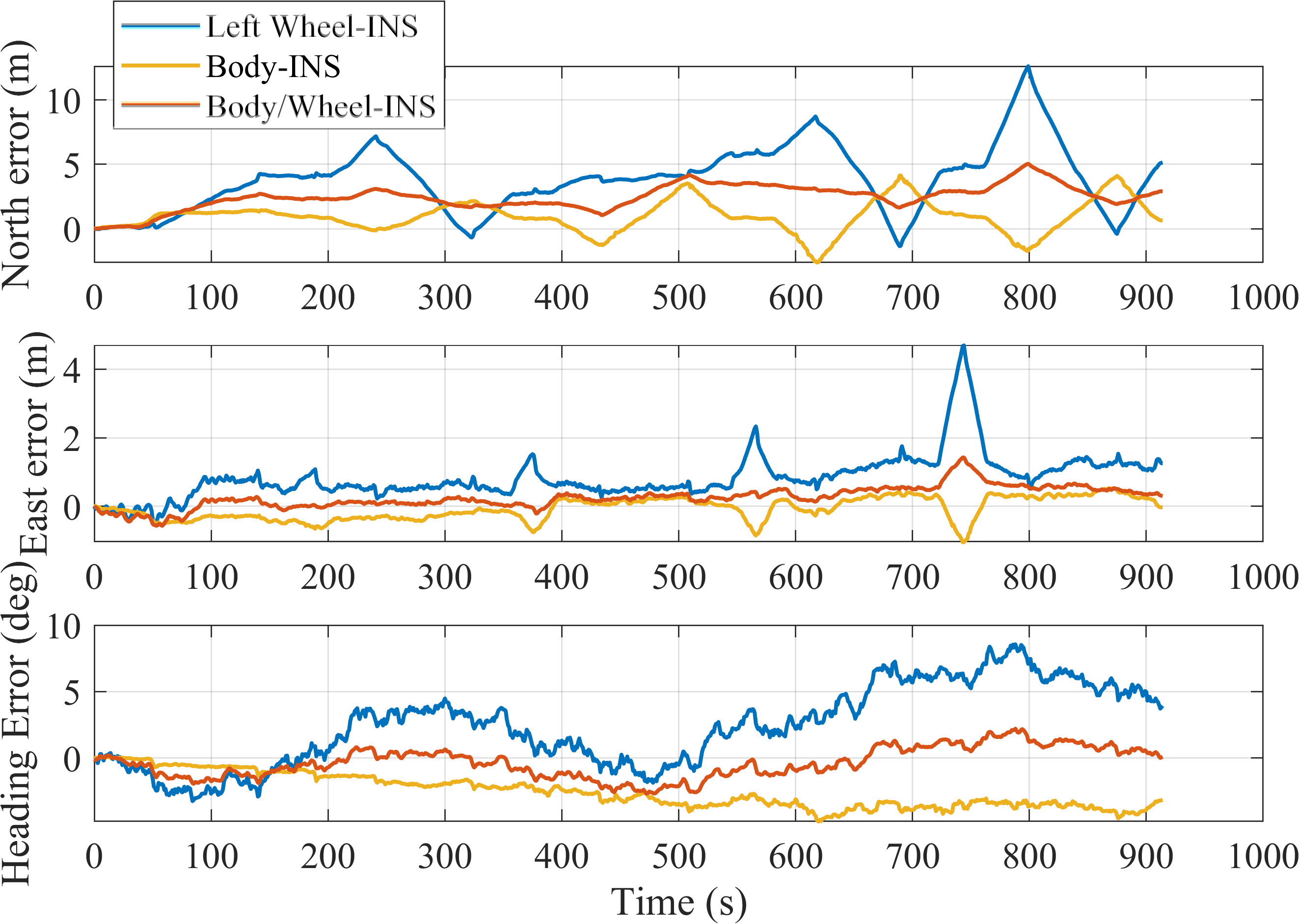}
	\caption{The horizontal positioning error and heading error of Body/Wheel-INS and the two subsystems.}
	\label{8}
\end{figure}

\begin{figure}[htbp]
	\centering
	\includegraphics[width=8.6cm]{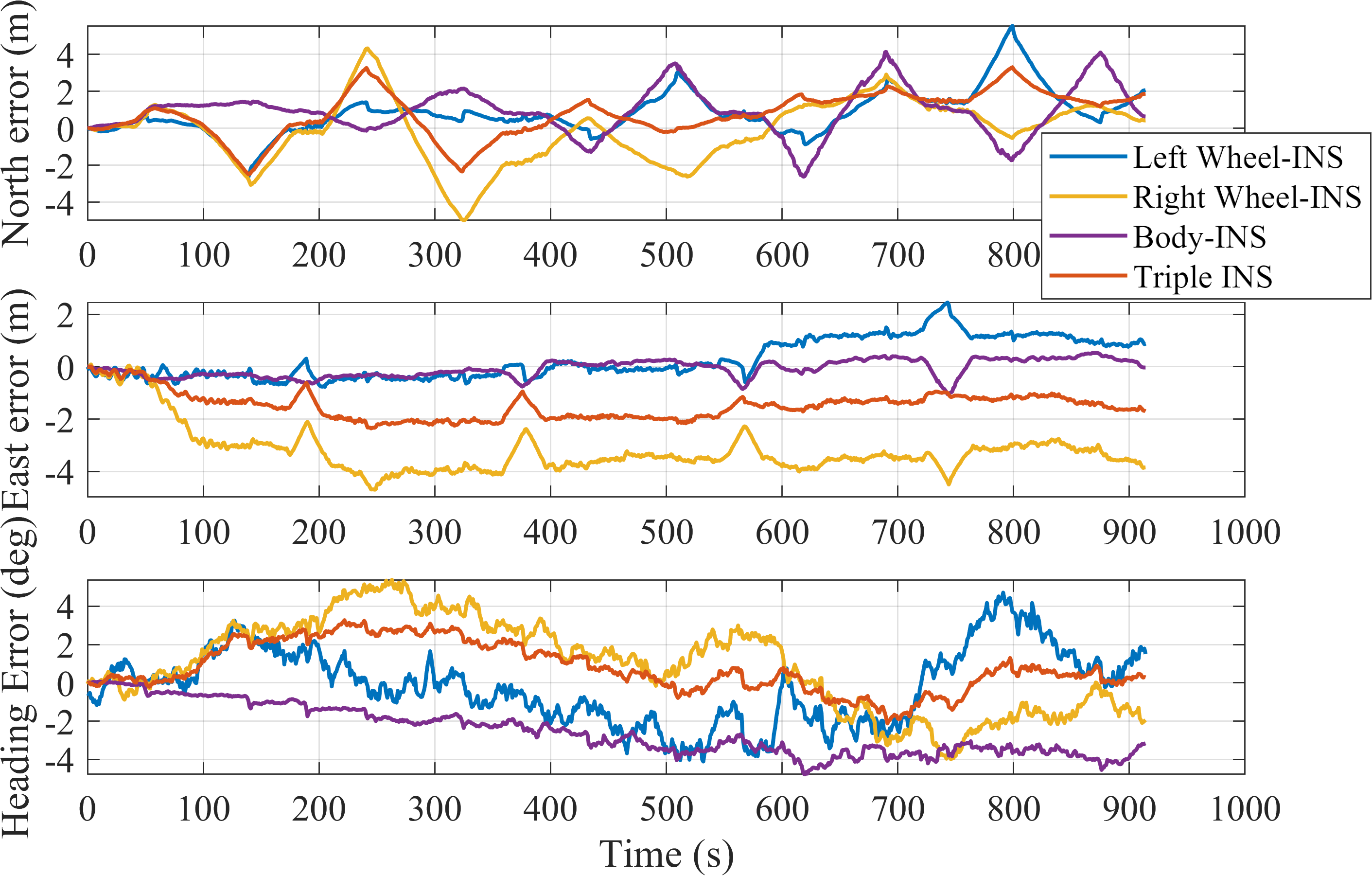}
	\caption{The horizontal positioning error and heading error of Triple INS and the three subsystems.}
	\label{9}
\end{figure}

Firstly, we compare the DR errors of the multi-IMU integrated system with that of the corresponding subsystems. The horizontal positioning errors and the heading errors of Dual Wheel-INS, Body/Wheel-INS, and Triple INS and the subsystems are pictured in Fig. \ref{7}, Fig. \ref{8}, and Fig. \ref{9}, respectively. In these experiments, the two inertial chips used in Dual Wheel-INS were C1 (left Wheel-IMU) and C4 (right Wheel-IMU). The two inertial chips used in Body/Wheel-INS were C2 (left Wheel-IMU) and C2 (Body-IMU). And the inertial chips used in Triple INS were C1 (left Wheel-IMU), C2 (right Wheel-IMU), and C4 (Body-IMU). In the three figures, ``Left Wheel-INS'' indicates Wheel-INS using the left Wheel-IMU; ``Right Wheel-INS'' indicates Wheel-INS using the right Wheel-IMU; ``Body-INS'' indicates the odometry-aided INS implemented using the Body-IMU and the wheel speed indicated by the Wheel-IMU.


Fig. \ref{7}-\ref{9} show that with the relative position constraint, the positioning and heading errors of the multiple IMUs-based DR systems (including Dual Wheel-INS, Body/Wheel-INS, and Triple INS) are situated between those of the corresponding subsystems. As explained in Section \uppercase\expandafter{\romannumeral5}-B, the position of the reference point is the combined result of the multiple IMU positions; thus the vehicle's position after measurement update would approach the middle of the estimated position of the multiple IMUs. In consequence, the DR error of the multi-IMU based DR system would also lay at the middle level between that of the subsystems.  

\begin{figure*}[htbp]
	\centering
	\subfigure[single Wheel-INS]{
		\includegraphics[width=8.5cm]{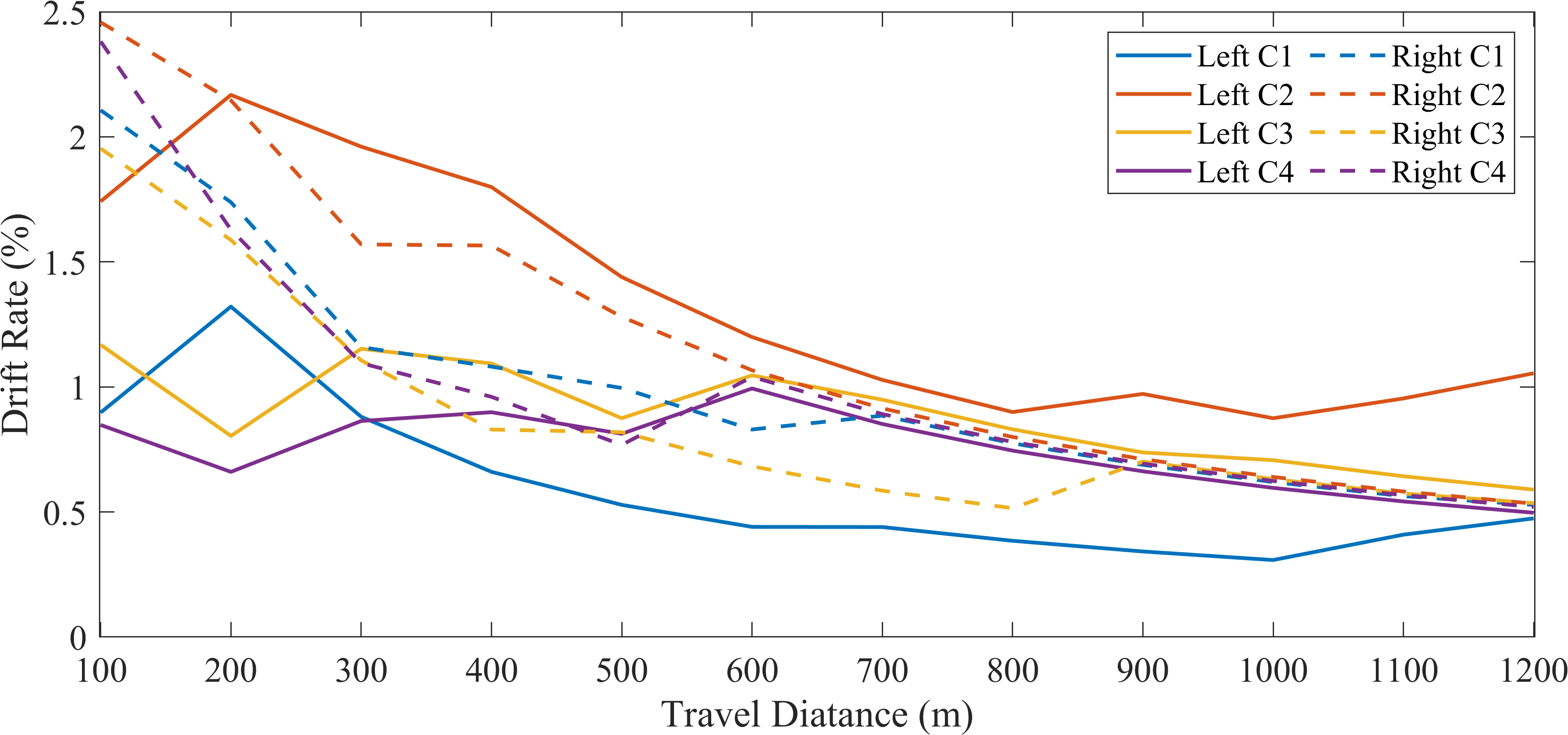}
	}
	\quad
	\subfigure[Dual Wheel-INS]{
		\includegraphics[width=8.5cm]{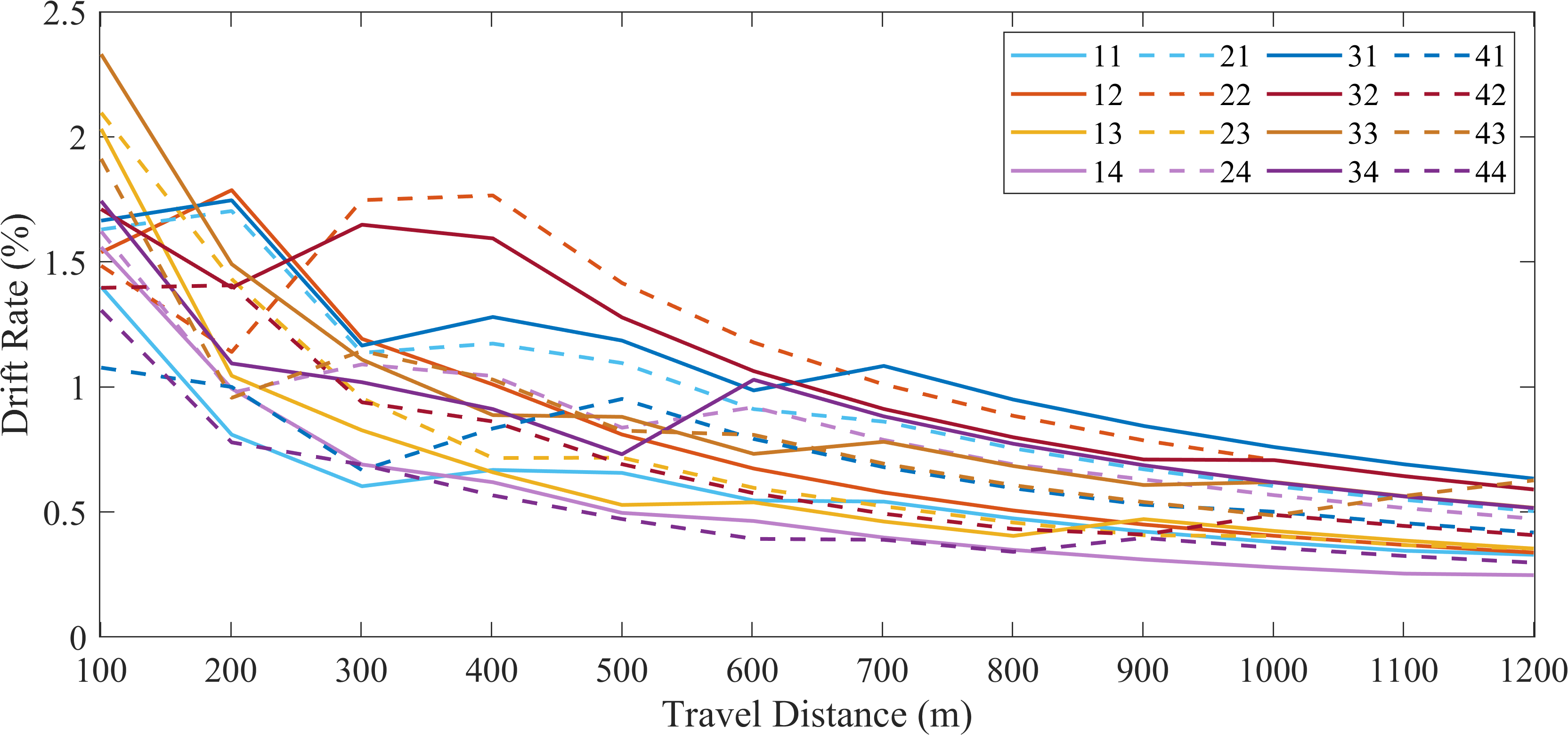}
	}
	\quad
	\subfigure[Body/Wheel-INS]{
		\includegraphics[width=8.5cm]{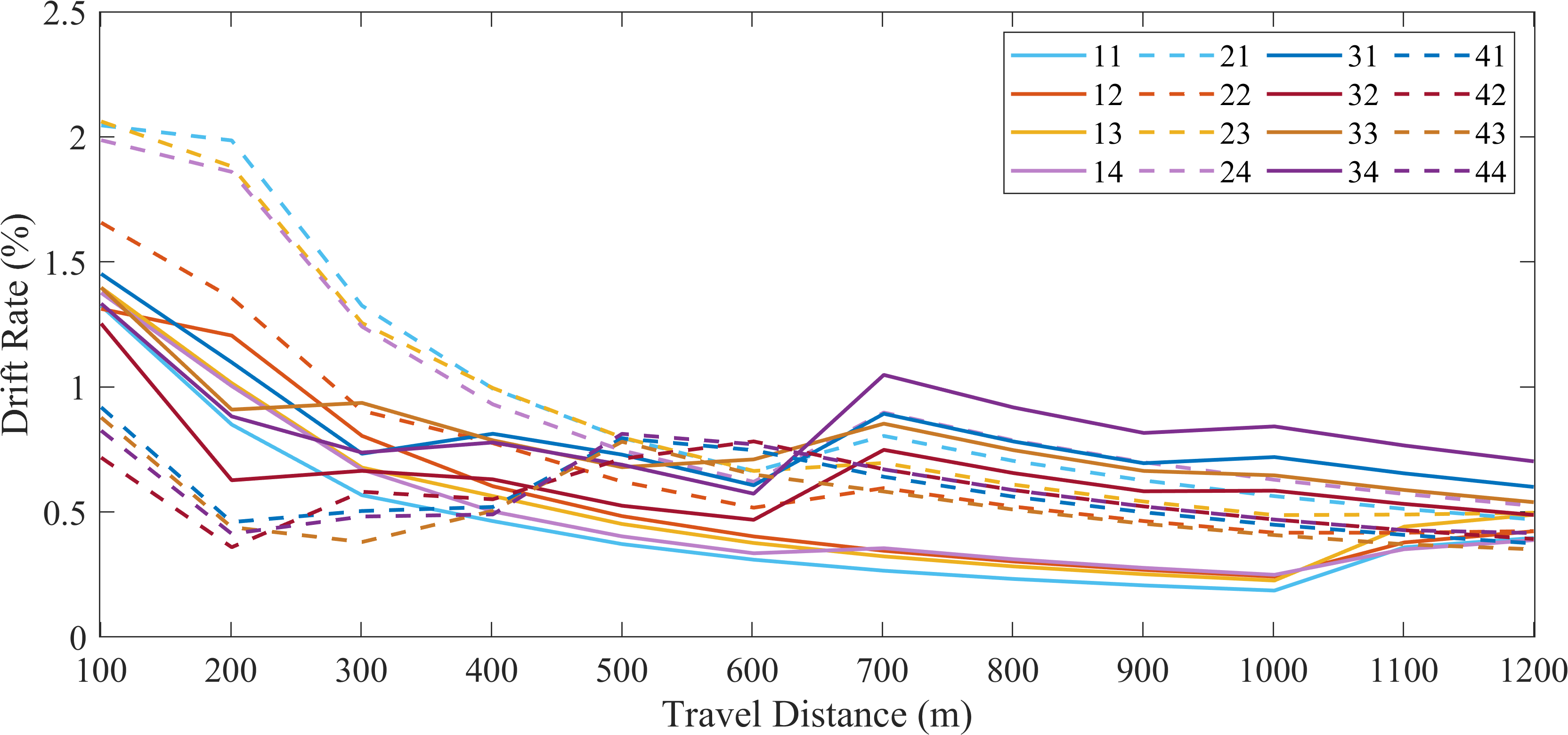}
	}
	\quad
	\subfigure[Triple INS]{
		\includegraphics[width=8.5cm]{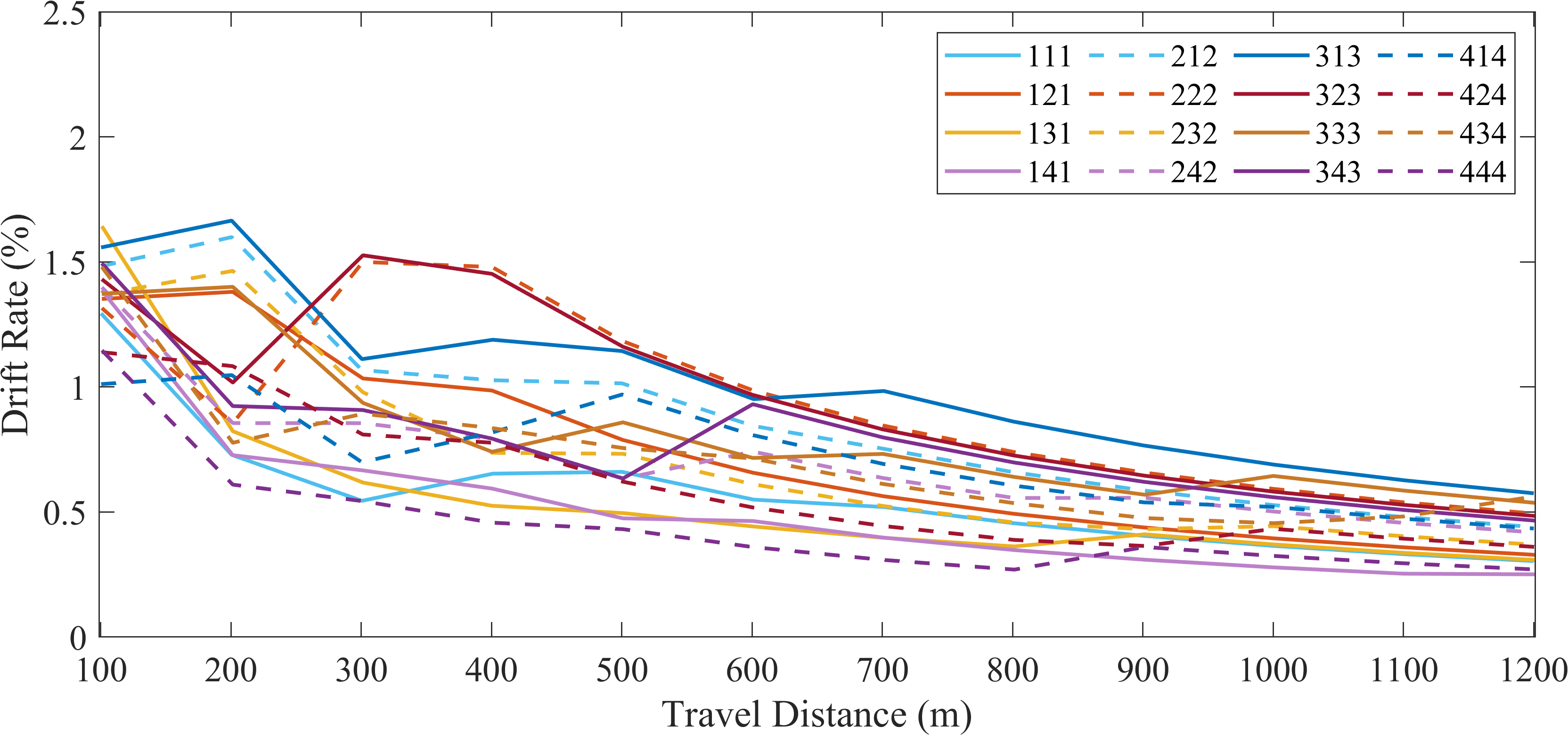}
	}
	\caption{Position drift rates of the four IMU configurations in all the experiments.}
	\label{10}
\end{figure*}

In our previous papers \cite{niu2019, wu2020arxiv}, we illustrated that calculating the misclosure error or the maximum position drift of the entire trajectory to evaluate the navigation performance of a DR system is not optimal. This is because the loop closure of the trajectory will suppress the error accumulation to some extent. And we adopted a different approach to demonstrate the localization performance. Firstly, we accumulated the traveled distance of the vehicle by a certain increment $( l )$ and calculated the maximum position error drift rate (= maximum position error/traveled distance) within each distance ( $l, 2l, 3l, ...$ ). Afterward, the mean value of the drift rates was computed as the final indicator. Here we use the same metric for the positioning performance evaluation. Concerning the evaluation of heading estimation performance, the root mean square error (RMSE) of the heading error was calculated. In this study, we chose $l$ as 100 m. The horizontal position drift rates of the four DR systems (single Wheel-INS, Dual Wheel-INS, Body/Wheel-INS, and Triple INS), which are functions of the traveled distance, are shown in Fig. \ref{10}. In Fig. \ref{10}(b), the two numbers of each curve represent the number of the inertial chips of the left and right Wheel-IMU used in Dual Wheel-INS, e.g., ``12" indicates that chip C1 of the left Wheel-IMU and chip C2 of the right Wheel-IMU were used. In Fig. \ref{10}(c), the two numbers of each curve represent the number of the inertial chips of the left Wheel-IMU and Body-IMU used in Body/Wheel-INS. Whereas in Triple INS (cf. Fig. \ref{10}(d)), the three numbers indicate the chip numbers of the left Wheel-IMU, right Wheel-IMU, and Body-IMU, respectively. Table \ref{Tab4} lists the error statistics of the eight sets of single Wheel-INS experiments. Table \ref{Tab5} lists the error statistics of the sixteen sets of Dual Wheel-INS experiments, sixteen sets of Body/Wheel-INS experiments, and sixteen sets of Triple-INS experiments. We can learn from Table \ref{Tab5} that Body/Wheel-INS exhibits the best position accuracy in the majority of cases (10 out of 16) and all the position drift rates in the sixteen experiments are less than 1\%.

It is evident in Fig. \ref{10} that with the increase of the traveled distance, the drift rate presents a downward trend. This is caused by the loop closure in the experimental trajectory which restrains the position drift. Additionally, it can be noted in Fig. \ref{10}(c) that the drift rate of ``21", ``22", and ``23" are very close in the initial phase and larger than other systems. This is because the positioning error of Wheel-INS using the chip C2 of the left Wheel-IMU is much larger than that of the other left Wheel-INS using different chips, which can be observed in Fig. \ref{10}(a). 


\begin{table}[h]
	\centering
	\caption{Error Statistics of the Single Wheel-INS}
	\label{Tab4}
	\begin{tabular}{m{1.2cm}<{\centering} m{0.48cm}<{\centering}p{0.48cm}<{\centering}p{0.48cm}<{\centering}p{0.48cm}<{\centering}p{0.48cm}<{\centering}p{0.48cm}<{\centering}p{0.48cm}<{\centering}p{0.48cm}<{\centering}}
		\toprule
		\multirow{3}{*}{Test} & \multicolumn{4}{c}{Left}  & \multicolumn{4}{c}{Right}\\
		\cmidrule(r){2-9}
	    &C1 &C2&C3 &C4&C1&C2&C3&C4\\	
		\midrule
		Drift Rate (\%) & 0.59&1.34&0.88&0.75&1.00&1.19&0.88&1.00\\
		Heading RMSE ($^\circ$) &1.93&3.82&2.62&4.44&2.69&2.52&4.41&2.28\\
		\bottomrule
	\end{tabular}
\end{table}

\begin{table}[h]
	\centering
	\caption{Error Statistics of the Dual Wheel-INS, Body/Wheel-INS and Triple INS}
	\label{Tab5}
	\begin{threeparttable}
	\begin{tabular}{p{0.7cm}<{\centering}p{0.7cm}<{\centering}p{0.7cm}<{\centering}p{0.7cm}<{\centering}p{0.7cm}<{\centering}p{0.7cm}<{\centering}p{0.7cm}<{\centering}}
		\toprule
		\multirow{3}{*}{Test} & \multicolumn{3}{c}{Position Drift Rate (\%)} & \multicolumn{3}{c}{Heading RMSE $(^\circ)$} \\
		\cmidrule(r){2-7}
		& DW\tnote{*} & BW\tnote{*}  & TW\tnote{*} & DW & BW & TW\\
		\midrule
		1&0.60& \textbf{0.46}& 0.57&2.12&\textbf{1.42}&2.11\\
		2&0.80& \textbf{0.56}& 0.73&1.62&1.58&\textbf{1.50}\\
		3&0.68& \textbf{0.54}& 0.56&2.86&\textbf{1.62}&2.77\\
		4&0.56& 0.52& \textbf{0.51}&0.95&1.49&\textbf{0.91}\\
		5&0.97& 0.96& \textbf{0.87}&\textbf{2.02}&2.13&2.24\\
		6&1.11& \textbf{0.72}& 0.93&2.33&\textbf{1.20}&1.97\\
		7&0.75& 0.92& \textbf{0.71}&3.55&\textbf{1.19}&4.02\\
		8&0.85& 0.96& \textbf{0.70}&2.13&2.07&\textbf{1.76}\\
		9&1.08& \textbf{0.82}& 1.01&2.81&3.44&\textbf{2.76}\\
		10&1.09& \textbf{0.65}& 0.95&2.46&2.39&\textbf{2.12}\\
		11&0.93& \textbf{0.79}& 0.81&4.37&\textbf{2.22}&4.79\\
		12&0.88& 0.84& \textbf{0.78}&1.46&3.99&\textbf{1.25}\\
		13&0.71& \textbf{0.57}& 0.72&\textbf{3.22}&4.04&3.54\\
		14&0.71& \textbf{0.56}& 0.61&2.19&3.74&\textbf{2.08}\\
		15&0.85& \textbf{0.53}& 0.72&4.27&\textbf{3.56}&4.51\\
		16&0.53& 0.57& \textbf{0.45}&1.66&4.41&\textbf{1.60}\\
		\bottomrule
	\end{tabular}
	\begin{tablenotes}   
	\footnotesize            
	\item[*]DW denotes Dual Wheel-INS; BW denotes Body/Wheel-INS; TW denotes Triple INS.   
	\end{tablenotes} 
\end{threeparttable}
\end{table}

Table \ref{Tab6} lists the mean value (MEAN) and STD (1$\sigma$) of the position drift rate and heading RMSE of the three multi-IMU-based DR systems in all the tests. It can be observed that benefiting from the multi-IMU position constraint, the position and heading accuracy of the three configurations have all been improved compared to the single Wheel-INS. The drift rates of Dual Wheel-INS and Body/Wheel-INS have been improved by approximately 14\% and 27\%, respectively. However, the positioning accuracy of Triple INS is similar to that of the Body/Wheel-INS (with a difference less than 0.05\%), instead of gaining further improvement. The heading accuracy of the three systems has been improved by about 19\%, 18\%, and 19\%, respectively. 

\begin{table}[h]
	\centering
	\caption{Error Statistics of All Four Kinds of IMU Configuration-based DR System}
	\label{Tab6}
	\begin{tabular}{ccccc}
		\toprule
		\multirow{3}{*}{IMU Configurations} & \multicolumn{2}{c}{Position Drift Rate (\%)} & \multicolumn{2}{c}{Heading Error $(^\circ)$} \\
		\cmidrule(r){2-5}
		& MEAN & STD & MEAN & STD \\
		\midrule
		Wheel-INS &0.95& 0.24&3.09& 0.99\\
		Dual Wheel-INS&0.82& 0.19&\textbf{2.50}& 0.97\\
		Body/Wheel-INS&\textbf{0.69}& 0.17&2.53& 1.14\\
		Triple INS&0.73& 0.16&2.50& 1.16\\
		\bottomrule
	\end{tabular}
\end{table}

\subsubsection{Comparison between the Distributed Filter and Centralized Filter}
To illustrate that the correlation issue in the proposed distributed filter system would not cause serious performance degradation while gaining significant computation efficiency, we implemented a centralized filter in Triple INS and compared it with the proposed distributed filter system. In the centralized filter, a long state vector (63 dimensions) containing the error-states of all three IMUs was constructed. Therefore, the correlation among the subsystems could be handled by the propagation of the covariance matrix. The measurement model built in the centralized filter was the same as that in the proposed distributed filter (cf. Section V-B). As described in Section-VI-B-1, we selected different chips in the three IMUs to get eight sets of experimental results. The navigation error statistics of the two systems in the experiments are listed in Table \ref{Tab7}. The algorithms were run on a laptop with an Intel i7-7700HQ CPU at 2.8 GHz for about one thousand times to get the average value. The mean processing time of the two main modules (velocity update and multi-IMU position constraint update) and the average running time of one IMU epoch in the two systems are compared in Table \ref{Tab8}.

It can be calculated from Table \ref{Tab7} that the mean position drift rate of the distributed filter and centralized filter in the eight experiments are 0.62\% and 0.59\%, respectively, while the mean heading RMSE are 1.70$^\circ$ and 1.78$^\circ$, respectively. We can conclude that the DR performance of the two filter systems is at the same level. Although the correlation terms between subsystems are modeled and handled well in the centralized filter, it only leads to a small fraction of performance improvement compared to the proposed distributed filter system. Because the MEMS IMUs used in our experiments were very cheap, they featured significant stochastic and instability errors which covered the potential accuracy increase from the proper mathematical model. Moreover, it can be observed from Table \ref{Tab8} that the proposed distributed filter runs about 8 times faster than the centralized filter. In the centralized filter, the matrix dimension would get larger with more IMUs or more states, which would dramatically increase the computation cost. However, in the distributed filter, the computation burden grows linearly with the increase of the IMU. The computation time can even be further reduced if the system is implemented via multi-thread. Therefore, the proposed distributed filter holds admirable advantages in terms of efficiency.   
\begin{table}[h]
	\centering
	\caption{Error Statistics of the Distributed Filter and Centralized Filter in Triple INS}
	\label{Tab7}
	\begin{threeparttable}
		\begin{tabular}{p{0.7cm}<{\centering}p{1.3cm}<{\centering}p{1.3cm}<{\centering}p{1.3cm}<{\centering}p{1.3cm}<{\centering}}
			\toprule
			\multirow{3}{*}{Test} & \multicolumn{2}{c}{Position Drift Rate (\%)} & \multicolumn{2}{c}{Heading RMSE $(^\circ)$} \\
			\cmidrule(r){2-5}
			& Distributed & Centralized  & Distributed & Centralized\\
			\midrule
			1&0.80& \textbf{0.63}& 2.14& \textbf{1.95}\\
			2&0.76& \textbf{0.63}& 1.71& \textbf{1.51}\\
			3&0.67& \textbf{0.64}& \textbf{2.96}& 3.93\\
			4&0.66& \textbf{0.55}& 1.26& \textbf{1.17}\\
			5&0.56& \textbf{0.53}& 1.94& \textbf{1.03}\\
			6&\textbf{0.67}& 0.72& \textbf{1.25}& 1.43\\
			7&\textbf{0.40}& 0.53& \textbf{1.76}& 2.49\\
			8&\textbf{0.46}& 0.52& \textbf{0.55}& 0.76\\
			
			\bottomrule
		\end{tabular}
		
	\end{threeparttable}
\end{table}

\begin{table}[h]
	\centering
	\caption{Running Time Statistics of the Distributed Filter and Centralized Filter in Triple INS}
	\label{Tab8}
	\begin{threeparttable}
		\begin{tabular}{p{0.85cm}<{\centering}p{0.85cm}<{\centering}p{0.85cm}<{\centering}p{0.85cm}<{\centering}p{0.85cm}<{\centering}p{0.85cm}<{\centering}}
			\toprule
			\multicolumn{2}{c}{Vel.\tnote{*} Update} & \multicolumn{2}{c}{Mul. Pos.\tnote{*} Update} & \multicolumn{2}{c}{Avg.\tnote{*} per Epoch}\\
			\cmidrule(r){1-6}
			 Dist.\tnote{*} & Cent.\tnote{*}  & Dist. & Cent. & Dist. & Cent.\\
			\midrule
			98& 694& 91& 690&120&1027\\
			\bottomrule
		\end{tabular}
		\begin{tablenotes}   
			\footnotesize            
			\item[*]Vel. denotes velocity; Mul. Pos. denotes multi-IMU position; Avg. denotes average; Dis. denotes distributed filter; Cent. denotes centralized filter. 
			\item[] Unit: $\mu$s. 
		\end{tablenotes}
	\end{threeparttable}
\end{table}

\subsection{Discussion}
By mounting two IMUs to the left and right wheel of the vehicle respectively, the wheel velocity information is doubled. As a result, the impact of the high-frequency noise can be reduced in Dual Wheel-INS. In addition, benefiting from the relative position constraint, the positioning robustness and reliability can be further improved compared with the single Wheel-INS. However, some errors in the subsystems are common-mode; thus, it is not sound to simply regard them as independent observations. In Body/Wheel-INS, the dynamic conditions of the two subsystems are significantly different thus there is a higher potential for performance improvement by information fusion. Although the constant gyro bias of the Body-IMU cannot be canceled as in the Wheel-IMU, its initial value is determined by the static data and remains stable over a short period (e.g., about 15 min in our experiment). Moreover, the NHC also helps mitigate the heading drift in Body-INS. Additionally, the dynamic condition at the vehicle body is lower than that of the wheel; thus some of the sensor errors such as the cross-coupling error will undermine the navigation performance of Body-INS more slightly than that of Wheel-INS. Furthermore, Body-INS shares the vehicle attitude with Wheel-INS in Body/Wheel-INS, enhancing the positioning performance to some extent. Therefore, Body/Wheel-INS somewhat outperforms Dual Wheel-INS in terms of positioning accuracy, while the heading accuracy of the two systems is comparable.

According to the construction procedure of the multi-IMU position constraint measurement, we can approximately regard the localization results of the multi-IMU integrated DR system as a weighted average of the subsystems. Although the weights of the subsystems can be set by the technical parameters of the inertial sensors, the real error level of each system cannot be reasonably determined. In particular, with the position constraint, the trajectories of the subsystems will be bundled together. In other words, if the position tracking results of the subsystems drift along contrary directions, the result of the multi-IMU integrated system will be better than that of all the subsystems. In contrast, if the positioning error of one subsystem is small while that of the other subsystem is relatively larger, the ultimate positioning accuracy of the integrated system will situate in the middle of that of the subsystems.

Additionally, although the correlation issue between the multi-IMU is not considered in the proposed system, it would not result in large accuracy deterioration compared to the centralized filter. This is because the significant stochastic errors in the low-cost inertial sensors dilute the contribution of the proper mathematical model. However, the proposed distributed filter exhibits desired computational efficiency with the decentralized structure.

We can also find that the positioning accuracy is not improved by adding another Wheel-IMU to Body/Wheel-INS, because the dynamic information obtained from one Body-IMU and one Wheel-IMU is sufficient to complement each other. That is to say, integrating a Body-IMU with Dual Wheel-INS can mitigate the positioning error to some extent while it is not the case by adding another Wheel-IMU to the Body/Wheel-INS. The reason is that there is a stronger complementarity between the Body-IMU and Wheel-IMU, while the dynamic information perceived by the two Wheel-IMUs is similar. In conclusion, it is not always the case that the positioning accuracy of the multi-IMU integrated system is superior to those of the subsystems. However, the reliability and robustness will be improved. 

Note that although we only show the experimental results with a wheeled robot moving at a low speed, the proposed multi-IMU system can be straightforwardly applied to cars. Experiments in \cite{niu2019} also show that Wheel-INS works with a car. However, when applying the multi-IMU system to a land vehicle with a high speed, one needs to make sure that the wheel rotation angular speed does not exceed the measuring range of the gyroscope of the Wheel-IMU.


\section{Conclusion}
In this article, the idea of fusing diverse dynamic information from different parts of the land vehicles to improve DR performance is investigated. Based on our previous study on Wheel-INS \cite{niu2019, wu2020arxiv}, a multiple MEMS IMUs-based DR system is proposed. It is necessary to mount at least one IMU on the non-steering wheel to obtain the vehicle velocity and take advantage of the rotation modulation. The multi-IMU DR system is implemented via a distributed EKF structure, where each IMU maintains its own state and updates it independently. The relative position constraint between the multiple IMUs is exploited by fusing their positions to obtain the coordinates of the reference point which is then treated as an external measurement in each EKF system. Particularly, three algorithms corresponding to three different multi-IMU configurations are implemented in this paper: Dual Wheel-INS, Body/Wheel-INS, and Triple INS.  

Field experiments demonstrate that both the positioning and heading accuracy of the three multi-IMU DR systems have been improved compared with the single Wheel-INS. In particular, Body/Wheel-INS outperforms Dual Wheel-INS in terms of position drift. Moreover, no performance improvement is obtained by Triple INS because one Wheel-IMU and one Body-IMU can complement each other sufficiently and more errors would be introduced if more low-cost sensors are integrated. Therefore, we recommend the Body/Wheel-INS configuration in consideration of computational burden and applicability. Some of the experimental data and code have been shared with the community (https://github.com/i2Nav-WHU/Wheel-INS).

By comparing the proposed system with the centralized filter in Triple INS, it can be learned that the distributed structure design would not result in obvious accuracy degradation while significantly reducing the computational loads.

The proposed multiple IMUs-based DR system can be considered as a type of wearable device for wheeled robots for autonomous ego-motion estimation. It can be installed directly without invading or modifying the vehicle's hardware. Additionally, the proposed solution can also be straightforwardly extended to more than three IMUs to form a sensor failure-resilient system with a trivial computational burden added. 

Future work includes investigating other information fusion algorithms to recover the weights of the subsystems more precisely and handle the error coupling issue between the subsystems more decently.

\begin{appendices}

\section{Estimator Parameters}
\label{EstimatorParameters}
The estimator parameters including the initial error state covariance and process noise covariance in the Kalman filter of the subsystems can be found in our open-sourced code. Particularly, the standard deviation of the vehicle velocity and multi-IMU position constraint measurement noise set in the experiments are listed in Table \ref{Tab9}, where wheel velocity contains both forward velocity and NHC.

\begin{table}[h]
	\renewcommand\arraystretch{1.5}
	\centering
	\caption{Standard Deviation of the Wheel Velocity and Multi-IMU Position Constraint Measurement Noise}
	\label{Tab9}
	\begin{threeparttable}
	\begin{tabular}{p{2.2cm}p{2.6cm}p{2.6cm}}
		\toprule
		\multicolumn{1}{c}{Measurement} & \multicolumn{1}{c}{Wheel-INS} & \multicolumn{1}{c}{Body-INS}\\
		\midrule
		{Wheel Velocity ($m/s$)} & diag${({0.03}, {0.02}, {0.02})}$ & diag${({0.05}, {0.04}, {0.04})}$
		\\
		{Multi-IMU Pos.\tnote{*} Constraint ($m$)} & diag${({0.2}, {0.2}, {0.2})}$ & diag${({0.2}, {0.2}, {0.2})}$
		\\
		\bottomrule
	\end{tabular}
	\begin{tablenotes}   
	\footnotesize            
	\item[*]Pos. denotes position.   
	\end{tablenotes}
\end{threeparttable}
\end{table}

Because the Wheel-IMU is directly mounted to the rear wheel of the vehicle, the forward velocity and NHC measurement in Wheel-INS would be more reliable. Consequently, we make the velocity constraint tighter in Wheel-INS than that in Body-INS by setting the standard deviation smaller in the filter. In addition, it is tricky to set the standard deviation of the multi-IMU position constraint measurement because it is highly related to the states of the multiple IMUs. If it is too small, the positions of the subsystems would become very close soon, while if it is too large, the constraint would be too loose to limit the error drift of the subsystems. Here we used the empirical value. However, the investigation of algorithms that determine the multi-IMU position constraint measurement noise adaptively would be more desirable.

\section{Design Matrices}
\label{DesignMatrices}
According to the velocity measurement model in Eq. \ref{velmeasurement} and the multi-IMU position constraint model in Eq. \ref{mltIMUmeasurement}, the corresponding design matrices (We choose left Wheel-INS as example here.) $\textbf{H}_v$ and $\textbf{H}_{mlt}$ are given by
\begin{equation}
\textbf{H}_v =
\begin{bmatrix}
\textbf{0}_{3} &\! {\textbf{C}}^v_n &\! \textbf{A} &\! \textbf{C}^v_n \textbf{C}^n_{b_1} \left( \bm{l}^{b_1}_{wheel} \times \right) &\! \textbf{0}_{3} &\! \textbf{D} &\! \textbf{0}_{3} \\
\end{bmatrix}
\end{equation}

\begin{equation}
\textbf{H}_{mlt} =
\begin{bmatrix}
\textbf{I}_{3} &\! \textbf{0}_3 &\! \left( \textbf{C}_v^{w_1}{\bm{r}}_{b_1}^{v} \right)\times &\! \textbf{0}_{3} &\! \textbf{0}_{3} &\! \textbf{0}_{3} &\! \textbf{0}_{3} \\
\end{bmatrix}
\end{equation}
where 
\begin{equation}
\begin{aligned}
\textbf{A} &= \begin{bmatrix} \textbf{0}_{3 \times 1} &\! \textbf{0}_{3 \times 1} &\! \textbf{B} \\ \end{bmatrix} + \textbf{C}^v_n \left[ \left(\textbf{C}^n_{b_1} \left( \bm{\omega}^{b_1}_{ib} \times \right) \bm{l}^{b_1}_{wheel} \right) \times \right] \\ 
\textbf{B} &= \left(- \textbf{C}^v_n \left[\left( {\bm{v}}^n_{b_1} \times \right) + \left({\textbf{C}}^n_{b_1} \left( \bm{\omega}^{b_1}_{ib} \times \right) \bm{l}^{b_1}_{wheel} \right) \times \right]\right){(:,3)} \\
\textbf{D} &= \textbf{C}^v_n \textbf{C}^n_{b_1} \left( \bm{l}^{b_1}_{wheel} \times \right)\mathrm{diag}(\bm{\omega}_{ib}^{b})\\
\end{aligned}
\end{equation} 
$\textbf{M}{(:,3)}$ represents the third column of matrix $\textbf{M}$; $\textbf{H}_v$ and $\textbf{H}_{mlt}$ represent the design matrix of the wheel velocity measurement and multi-IMU position constraint measurement, respectively; $\textbf{A}$ contains the vehicle heading error indicated by Wheel-IMU and the attitude error of the Wheel-IMU.

\end{appendices}

\ifCLASSOPTIONcaptionsoff
  \newpage
\fi

{
\small
\bibliographystyle{IEEEtranTIE}
\bibliography{ReferenceWheel-INS3}
}

\begin{IEEEbiography}[{\includegraphics[width=1in,height=1.25in,clip,keepaspectratio]{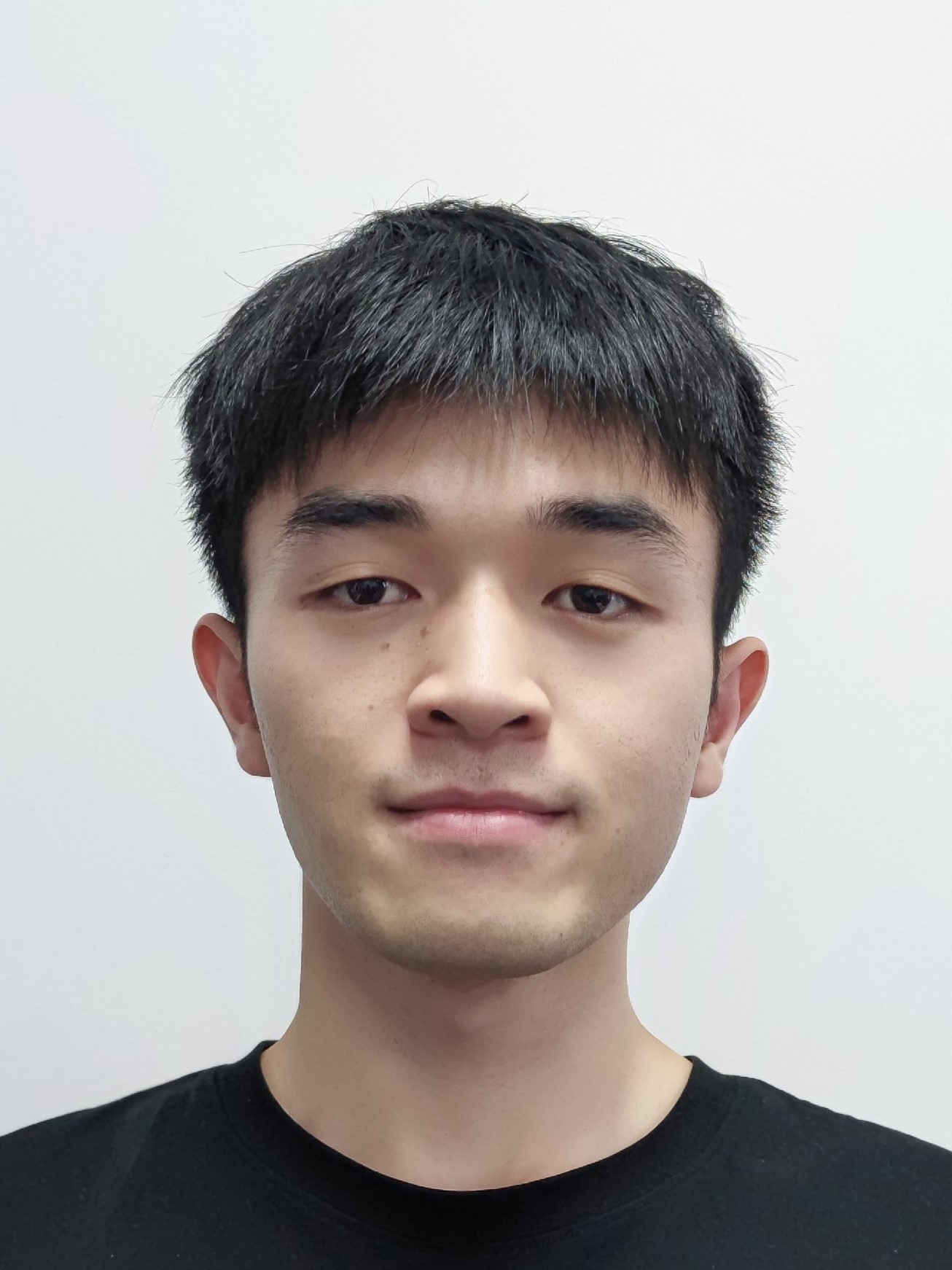}}]{Yibin Wu} received the B.Eng. degree (with honors) in Navigation Engineering and the M. Eng. degree in Navigation, Guidance and Control from Wuhan University, Wuhan, China, in 2017 and 2020, respectively. His research interests focus on inertial navigation, multi-sensor fusion, and mobile robot state estimation.
\end{IEEEbiography}
\begin{IEEEbiography}[{\includegraphics[width=1in,height=1.25in,clip,keepaspectratio]{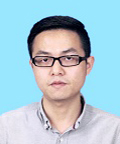}}]{Jian Kuang} received the B.Eng. degree and Ph.D. degree in Geodesy and Survey Engineering from Wuhan University, Wuhan, China, in 2013 and 2019, respectively. He is currently a Postdoctoral Fellow with the GNSS Research Center in Wuhan University, Wuhan, China. His research interests focus on inertial navigation, pedestrian navigation and indoor positioning.
\end{IEEEbiography}
\begin{IEEEbiography}[{\includegraphics[width=1in,height=1.25in,clip,keepaspectratio]{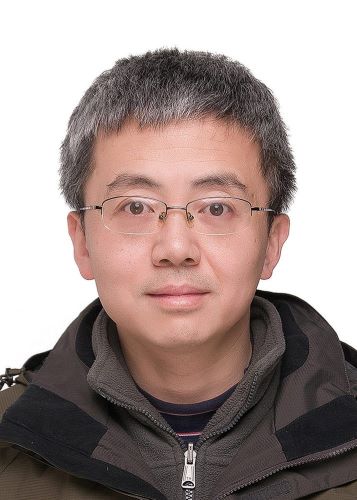}}]{Xiaoji Niu} received the B.Eng. degree (with honors) in Mechanical Engineering and the Ph.D. from Tsinghua University, Beijing, China, in 1997 and 2002, respectively. From 2003 to 2007, he was a Post-Doctoral Fellow with the Department of Geomatics Engineering, University of Calgary. From 2007 to 2009, he was a senior scientist with SiRF Technology, Inc. He is currently a Professor of the GNSS Research Center and the Artificial Intelligence Institute of Wuhan University, Wuhan, China. His research interests focus on INS, GNSS/INS integration for land vehicle navigation and pedestrian navigation.
\end{IEEEbiography}


\end{document}